\documentclass[11pt]{article}

\PassOptionsToPackage{hyperfootnotes=false}{hyperref}
\usepackage{silence}

\usepackage[final]{acl}
\usepackage{times}
\usepackage{latexsym}
\usepackage{booktabs}
\usepackage{xcolor}
\usepackage{enumitem}
\usepackage{amsmath}
\usepackage{makecell}
\usepackage[T1]{fontenc}

\usepackage[utf8]{inputenc}

\usepackage{microtype}

\usepackage{inconsolata}

\usepackage{graphicx}
\usepackage{tabularx}
\usepackage{multirow}
\usepackage{placeins}
\usepackage{tcolorbox}

\title{ChildEval:WHEN LARGE LANGUAGE MODELS MEET
CHILDREN’S PERSONALITIES}

\author{
  \textbf{Yanyan Luo},
  \textbf{Xue Han\textsuperscript{*}},
  \textbf{Chunxu Zhao},
  \textbf{Ruiqiao Bai},
  \textbf{Yaxing Zhang},
  \\
  \textbf{Qian Hu},
  \textbf{Lijun Mei},
  \textbf{Junlan Feng\thanks{The corresponding author.}}
\\
   JIUTIAN Research, China Mobile, Beijing, China,
\\
  \small{
    \href{mailto:email@domain}{\{luoyanyan,hanxue,zhaochunxu,bairuiqiao,zhangyaxing,huqian,meilijun,fengjunlan\}it@chinamobile.com}
  }
}

\begin{document}
\maketitle
\begin{abstract}
While LLMs enable personalized chatbots, their effectiveness in child-centered personalization remains unclear, as systematic evaluation of child-specific preferences is still lacking. To address this gap, we introduce ChildEval, a benchmark for evaluating LLMs’ ability to infer and follow child-centered preferences in long-context conversations. ChildEval contains 29K synthesized persona profiles of children aged 3–6, providing relatively static background information. Each persona is associated with a child preference—which may align with, conflict with, or be independent of the persona—expressed either explicitly in a single sentence or implicitly through 6–10 turn dialogues. Explicit and implicit preferences are designed to reflect the same underlying preference but differ in expression, capturing dynamic aspects of preference expression rather than changes in the static persona. The benchmark spans five top-level and fourteen sub-level categories covering children’s daily lives and development. We further propose fine-grained, child-centric evaluation protocols to systematically assess open-source LLMs. Experimental results demonstrate how different personalized representations affect LLM responses and suggest that finetuning on ChildEval can enhance child-centered performance. Our code and dataset are available at \url{https://github.com/ziyanluo/ChildEval}.
\end{abstract}

\section{Introduction}
Large Language Models (LLMs) (e.g., ChatGPT \citep{openai2024openaio1card}, Gemini \citep{geminiteam2025geminifamilyhighlycapable}, and Claude \citep{TheC3}) have achieved remarkable success in effectively understanding and generating human language, leading to a revolutionary era in LLMs \citep{xu-etal-2025-personalized,ICLR2025_03945380}. Beyond generic dialogues, LLMs have been utilized in a wide range of individual daily tasks (e.g., healthcare \citep{xu2024talk2care,HAN202389}, finance \citep{easin2024intelligent} and other domains \citep{bai-etal-2025-self,wang-etal-2025-multipl}) to deliver personalized user experiences based on preferences \citep{kumar2024longlampbenchmarkpersonalizedlongform}. 
One promising direction is to provide personalized responses tailored to children’s preferences, fostering their development and learning while supporting caregivers \cite{feng2024genericllmshelpanalyze,Seo_2024,Seo_2024_2}, rather than merely returning generic correct answers.

Previous research on LLM personalization has mainly focused on adult preferences and tasks.
\citet{qiu2025measuringmakesuniquedifferenceaware} advance personalization by extracting user preferences from historical profiles, while other studies address the lack of benchmarks for evaluating personalized capabilities \citep{salemi2024lamplargelanguagemodels, jiang2025knowmerespondme}.
However, the proposed benchmarks focus on general preferences (e.g., the number of dialogue turns) with generic tasks (e.g., ticket booking and restaurant recommendations) for adults. Several benchmarks target children \citep{rath2025childrenharms, liu2024benchmarking}, but with limited scope: \citet{rath2025childrenharms} focus on child safety without diverse child-centered tasks, and \citet{liu2024benchmarking} examine style imitation without considering developmental or learning needs. Thus, whether LLMs can deliver personalized responses aligned with child-specific preferences remains underexplored.


In particular, we identify two key gaps in existing child personalization benchmarks for LLMs.
(1) Current benchmarks fail to capture children’s distinct preferences and developmental needs\citep{liu2024benchmarking}, 
limiting age-appropriate personalization \citep{stakeholder2024}. 
(2) There is no comprehensive evaluation taxonomy tailored to child personalization: existing studies largely focus on adult-centered preferences (e.g., general preference following) and overlook children’s specific needs, such as generating developmentally appropriate content and supporting creativity.


Facing these challenges, we introduce ChildEval, a benchmark for evaluating LLMs' ability to infer, interpret, and follow child-centered preferences in long-context conversations. Targeting preschool children (ages 3–6) with high demand for LLM-based chatbot companionship, ChildEval comprises 29K synthesized children’s persona profiles, with explicit preferences aligned to each persona and their implicit expressions revealed through multi-turn dialogues consisting of 6 to 10 turns.
The preferences cover 5 top-level and 14 sub-level topics that involve children's daily lives and development \citep{antle2008child,PingWang2013} according to the guidelines published by the Ministry of Education of China \citep{MOE2012Guidelines}. We also propose fine-grained, child-centric evaluation protocols to systematically assess open-source LLMs. Experimental results demonstrate the influence of personalized representations on model responses and suggest that finetuning on this dataset can improve performance.
\section{Related Work}
\textbf{Personalization Data and Evaluation.} PersonaMem \citep{jiang2025knowmerespondme} and HiCUPID \citep{kim2024hicupid} simulate multi-attribute personas and multi-turn interactions to assess user-specific consistency. Evaluation frameworks consider metrics such as style alignment, preference fidelity, and user satisfaction \citep{salemi2024lamplargelanguagemodels}. To address the scarcity of personalized data, prior work explores synthetic dialogue generation \citep{braga2024syntheticqa}, profile summarization \citep{zhang2024guidedpersona}, and memory retrieval from past interactions \citep{liu2023memprompt}.

Child-centered personalization has recently gained attention. KidLM \citep{nayeem2024kidlm} introduces child-oriented corpora and training strategies, complemented by work on style simplification \citep{valentini2023automaticgenerationsimplificationchildrens} and child-specific safety evaluation \citep{rath2025childrenharms}. However, existing benchmarks remain largely general-purpose, leaving it unclear whether LLMs can follow children’s preferences in multi-turn settings.

\textbf{LLM Personalization Methods.}
Methodological advances in personalization broadly fall into non-parametric and parametric approaches. Non-parametric methods condition model behavior without modifying parameters, including \emph{prompt-based} personalization using explicit or inferred traits \citep{zhang2024personalized,liu2024once,wu2024profile}, \emph{memory-based} approaches that leverage past interactions \citep{madaan2022memprompt,zhang2024malp,magister2024plum,zhang2025prime}, and \emph{preference inference and conditioning} that extracts preferences to guide generation \citep{zhao2025prefeval,ryan2025synthesizeme,adila2025chameleon}.
Parametric approaches embed user traits into model parameters via finetuning or reinforcement learning, including parameter-efficient adaptations and reward-based personalization \citep{wang2025oppu,castricato2024e2p,zhuang2024hydra,li2024prlhf,zhao2025rlpa}.
Despite progress, it remains unclear whether these methods generalize to children, motivating child-specific evaluation.
\begin{figure*}[ht]
\begin{center}

\includegraphics[width=0.65\linewidth]
{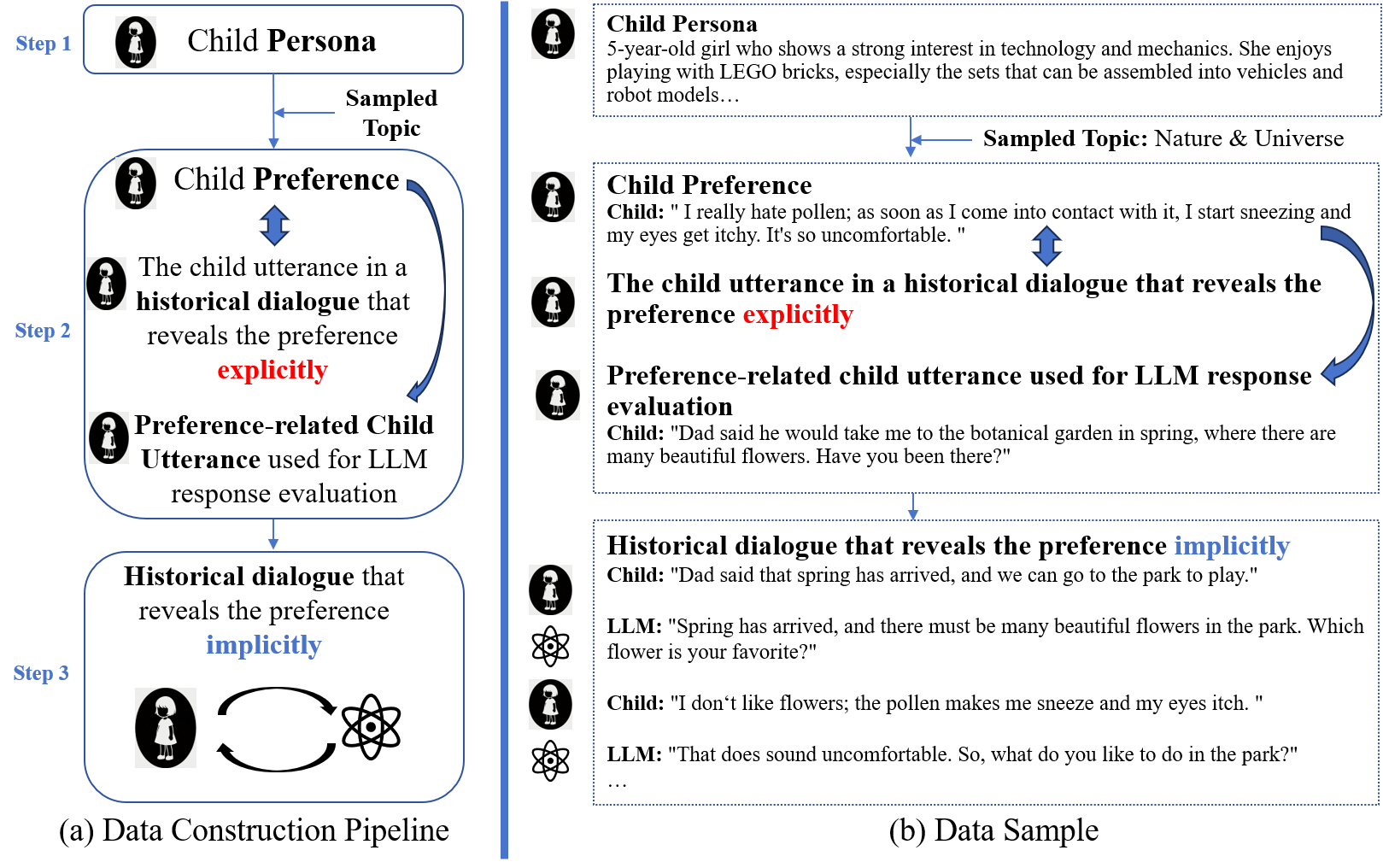} 
\end{center}
\caption{Overview of the ChildEval benchmark.(a) Data Construction Pipeline.(b) A data sample includes a child persona, a child preference statement, which is identical to the child utterance in the historical dialogue that explicitly reveals the preference, a historical multi-turn dialogues that implicitly reveal the preference, and a piece of preference-related child utterance used for LLM response evaluation.
}
\label{fig:overview}
\end{figure*}

\section{ChildEval Benchmark}
\subsection{Problem Formulation}
To evaluate whether an LLM can perceive and adapt to a child's preference $\rho$ when it communicates with the child, the full prompt sent to the model could be formulated by:

\begin{equation}
\begin{aligned}
  \mathcal{B} = H + u^*
\end{aligned}
\end{equation}

where 
\begin{itemize}[ itemsep=2pt, topsep=2pt]
\item $+$ denotes the concatenation of texts.
    \item $H = \{ S_1, S_2, \dots, S_t, \dots, S_T \}$ denotes a multi-session conversation history between a child and an LLM. Each session $S_t = \{ (u_{t,1}, m_{t,1}), \dots, (u_{t,K_t}, m_{t,K_t}) \}$ consists of $K_t$ dialogue turns, where $u_{t,k_t}$ is the child utterance and $m_{t,k_t}$ the model response.  
    \item $u^*$ is a child utterance related to the child preference $\rho$, and would be used as a utterance that the LLM shall respond to.
\end{itemize}

Sessions in $H$ are categorized as: 
\begin{itemize}[ itemsep=2pt, topsep=2pt]
    \item \textbf{Relevant session}: Following the setting of \citet{zhao2025prefeval}, the first session $S_1$ of $H$ is a session with dialogues related to the user preference queried by $u^*$.
    
    \item \textbf{Irrelevant session}: The remaining sessions of $H$ contain dialogue turns unrelated to $u^*$.

\end{itemize}
In each relevant session, the child preference $\rho$ can be revealed explicitly or implicitly:
\begin{itemize}[itemsep=2pt, topsep=2pt]
    
    \item \textbf{Explicit}: Such a session contains a single dialogue turn  $S_1 = \{ (u_{1,1}, m_{1,1})\}$, and $u_{1,1}$ directly expresses the child preference.
    \item \textbf{Implicit}: Such a session contains multiple dialogue turns, and the user preference could be implicitly inferred by partial user dialogue turns in this session.
\end{itemize}

The task used for evaluating the LLM could then be formulated as:  

\begin{equation}
\begin{aligned}
    f(p,\mathcal{B}) \;\longrightarrow\; \hat{m}
\end{aligned}
\end{equation}

where $p$ denotes the child persona, i.e. the persistent attributes that embody the child's consistent personality traits (e.g., age and gender) and long-standing interests. $f(.)$ denotes the model to be evaluated, and $\hat{m}$ is the response generated by the model given the prompt $\mathcal{B}$. A good response shall align with the child persona $p$ and the child's preference $\rho$ revealed by the historical conversation displaced in $\mathcal{B}$. 

Figure \ref{fig:overview}(b) shows a ChildEval example, which includes a child persona, explicit and implicit preference cues from historical dialogues, and a preference-related utterance for evaluation, enabling assessment of preference adaptation and persona effects.


\begin{table*}[htb]
\small
\begin{center}
\resizebox{0.8\textwidth}{!}{\begin{tabular}{l|llll}
\toprule
\bf Topic & \multicolumn{4}{l}{\bf Subtopic} 
\\ \hline
Art enlightenment & Music & Dance & Painting \& Crafts & \\ 
(21.64\%) & (6.68\%) & (6.74\%) & (8.22\%) & \\
Cognitive development and exploration & Science & Nature \& universe & Technology & Learning\\ 
(29.20\%) & (7.72\%) & (7.19\%) & (7.09\%) & (7.20\%)\\
Nutrition and physical activity & Outdoor activity  & Health eating& & \\ 
(13.11\%) & (6.24\%) & (6.87\%) &  & \\
Language and literacy development & Story & Language & Reading & \\ 
(22.09\%) & (7.39\%) & (7.31\%) & (7.39\%) & \\
Social and emotional development & Social interaction & Play &  & \\ 
(13.97\%) & (6.94\%) & (7.03\%) &  & \\
\bottomrule
\end{tabular}
}
\end{center}
\caption{Distribution of the 14 preference topics within ChildEval, which are related to children's daily life and development.}
\label{topic-table}
\end{table*}

\subsection{Data Construction Pipeline}
The data construction pipeline comprises three steps, as shown in Figure~\ref{fig:overview}(a).

\textbf{Step 1}: We generate 29K child personas using Qwen2.5-72B via iterative generation and refinement. Semantic duplicates are removed with FAISS \citep{douze2024faiss} based on text embeddings \citep{bge_embedding} to ensure diversity. All child names and identifiers are eliminated through LLM processing and human review to mitigate privacy risks.

\textbf{Step 2}: For each child persona, we generate two explicit preferences using persona-conditioned prompts. Each preference may align with, conflict with, or be independent of the persona, reflecting context-dependent expression rather than changes in the static persona. Preferences are expressed as single first-person sentences, each grounded in a distinct sub-level topic from a predefined taxonomy (Table~\ref{topic-table}) covering children’s daily lives and development \citep{antle2008child, PingWang2013, MOE2012Guidelines}, and generated together with a corresponding child-style utterance to initiate dialogue. This process yields 58K explicit preferences, of which 46K are retained after FAISS-based semantic filtering. 


\textbf{Step 3}: 
Historical dialogues that implicitly reveal preferences are generated via prompt-based generation conditioned on the corresponding explicit preference, with self-verification enforcing predefined rules (Section~\ref{prompts}), resulting in 6–10-turn child–LLM conversations.
These implicit preferences convey the same underlying preference as their corresponding explicit ones, but differ in expression.



\subsection{Fine-grained Evaluation Metrics for Child Preferences}
Existing personalization evaluation studies mainly cover standard adult-centered preferences and fail to capture the unique requirements of child-oriented interactions \citep{zhao2025prefeval, jiang2025knowmerespondme}. To address this gap, we propose fine-grained evaluation metrics tailored for children's personalization.


\textbf{(1) Preference Consistency (PC).} 
We evaluate whether the model’s responses align with a child’s explicitly stated or implicitly inferred preferences, using established criteria from prior work \citep{zhao2025prefeval}, which are released under the Creative Commons Attribution-NonCommercial 4.0 International Public License and adapted to child-oriented dialogues.

\textbf{(2) Child-Oriented Evaluation (COE).}
Beyond preference consistency, child-centered dialogues require additional fine-grained evaluation dimensions that differ from typical adult-oriented communication. Accordingly, we propose a set of novel child-oriented evaluation metrics focusing on the distinctive linguistic and contextual characteristics of child-centered conversations, covering four sub-dimensions.

\textbf{\emph{Emotional Adaptation (EA).}} The LLMs should be sensitive to the emotions expressed by the children, providing empathetic, supportive, and age-appropriate responses that help to maintain a positive atmosphere of interaction.  

\textbf{\emph{Interaction Scaffolding (IS).}} The LLM should be able to scaffold effective child-centered conversation with prompts, clarifications, or playful follow-ups in a natural conversational flow.

\textbf{\emph{Developmental Appropriateness (DA).}} The LLM's responses should match the cognitive and linguistic abilities of 3-to-6-year-old children, avoiding overly complex vocabulary or reasoning while providing informative and stimulating content. 

\textbf{\emph{Engagement (EG).}} The LLM should be able to produce lively and appealing utterances, using child-specific markers such as playful particles, reduplication, or culturally grounded scenarios, to keep children actively interested in the dialogue.

\section{Experiments and Analyses} 

\begin{figure*}
\begin{center}
\includegraphics[scale=0.24]
{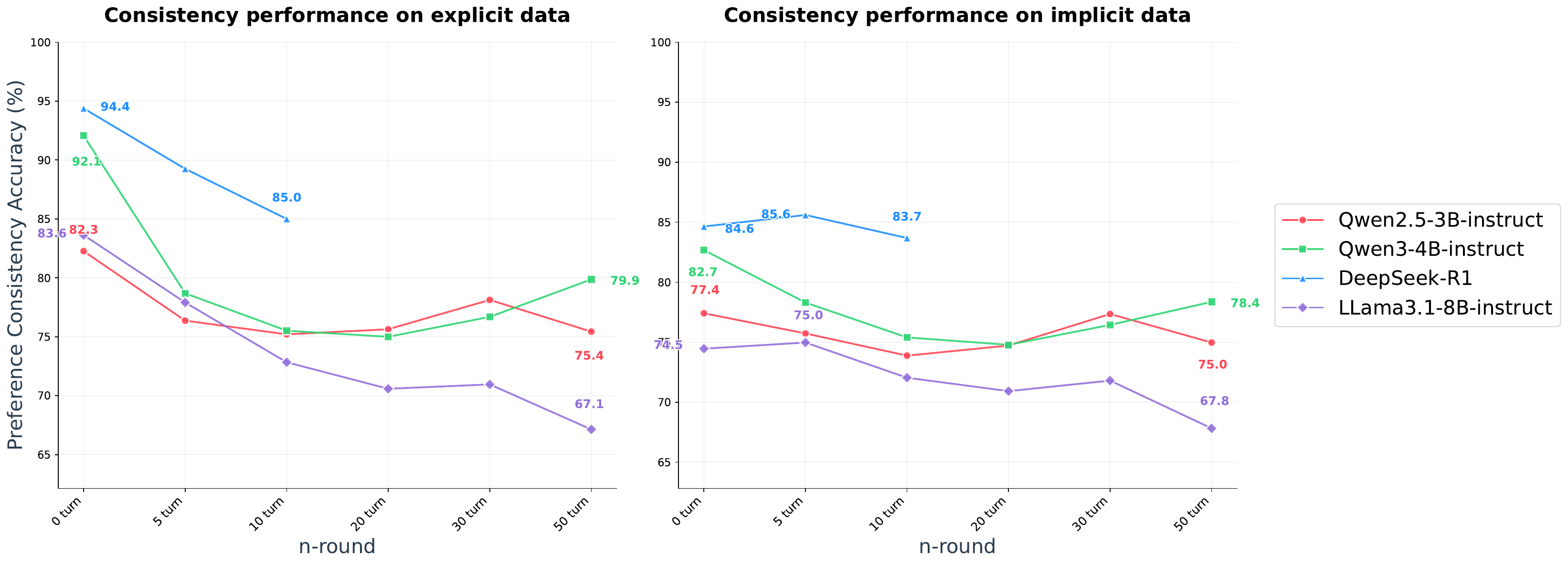}
\end{center}

\caption{Zero-shot consistency of LLMs with children’s explicit (left) and implicit (right) preferences across n-turn dialogues. Each n-turn dialogue uses a fixed token set (See Table ~\ref{table:token}). Due to the 8,192-token context limit enforced by the DeepSeek-R1 API, evaluations for this model are restricted to at most 10 turns.}
\label{fig:preference}
\end{figure*}

\subsection{Experimental Setup}
We evaluate five open-source LLMs (Qwen2.5-3B/72B-Instruct, Qwen3-4B-Instruct, LLaMA3.1-8B-Instruct, DeepSeek-R1-671B, and Mistral-7B-Instruct-v0.3) using three child-preference adaptation strategies: prompt-based (PBM), LoRA finetuning, and our Persona Steer Model (PSM), which injects child persona information from ChildEval via a pluggable Persona Steer Module (Section~\ref{sec:steer}). All experiments are zero-shot on a bilingual (Chinese–English) dataset, with additional long-context multi-session inputs to assess preference modeling under extended context (Appendix~\ref{Long_Context_Settings}). Qwen2.5-3B-Instruct serves as the SFT backbone, Qwen2.5-72B-Instruct is used for evaluation, and ChildEval is split 8:2 into train/test sets for a total of 46K samples. Fine-tuning and evaluation were conducted on two and four A800 GPUs, respectively.

\subsection{Reliability of LLM-as-a-Judge}
\label{sec:llm-judge}
To assess the reliability of machine-based evaluation, we conducted a human evaluation on a subset of the data and examined its consistency with LLM-based judgments. Specifically, we randomly sampled 600 examples covering all preference types and manually annotated them, then computed the agreement rates between human judgments and LLM evaluations. As shown in Table~\ref{tab:consistency_child}, the agreement rate reaches 88.83\% for PC. While LLM-based evaluation cannot fully substitute for human review, this result indicates that it provides a reliable and scalable mechanism for assessing preference consistency.

\begin{table}[htbp]
  \centering
  \small 
  \setlength{\tabcolsep}{3pt}
  \resizebox{0.92\linewidth}{!}{
  \begin{tabular}{l c c c}
    \toprule
    \multicolumn{2}{c}{\textbf{PC}} & \multicolumn{2}{c}{\textbf{COE}} \\
    \cmidrule(lr){1-2} \cmidrule(lr){3-4}
    \textbf{Sub-item} & \textbf{Rate (\%)} & \textbf{Dimension} & \textbf{Rate (\%)} \\
    \midrule
     Acknowledge Preference  & 78.17 & EA & 99.50 \\
     Violate Preference      & 89.67& IS & 99.33 \\
     Hallucinate Preference  & 62.00 & DA & 99.50 \\
    Helpful Response        & 98.83 & EG & 99.33 \\
    \midrule
    PC & 88.83 & \multicolumn{2}{c}{-} \\
    \bottomrule
  \end{tabular}
  }
  \caption{Human-verified agreement on Preference Consistency (PC), following established evaluation criteria \citep{zhao2025prefeval}, as well as on Child-Oriented Evaluation (COE).}
  \label{tab:consistency_child}
\end{table}


The high agreement rates on COE criteria in Table~\ref{tab:consistency_child} indicate that these criteria are well-defined and consistently applicable across different evaluators. This consistency suggests that the proposed criteria effectively capture the core requirements of child-directed communication and can be reliably applied at scale. Rather than distinguishing subtle stylistic variations, the COE focuses on assessing whether model responses satisfy essential principles of child-directed interaction under a unified safety standard. This design ensures that the evaluation remains robust, reproducible, and well aligned with the practical objectives of child-oriented model assessment.

\subsection{Cross-Model Bias Analysis for Validating LLM-as-a-Judge}

To further assess potential model-specific bias and validate our evaluation metrics, we use the same randomly sampled examples as in Section~\ref{sec:llm-judge}, covering all preference types, and evaluate them using DeepSeek-R1. We compute agreement rates between DeepSeek-R1 and Qwen2.5-72B, and additionally compare the outputs of both models against human judgments. As shown in Tables~\ref{tab:consistency_child} and~\ref{tab:deepseek_agreement}, the evaluations produced by the two models are highly consistent and closely aligned with human assessments. Notably, for child-oriented evaluation metrics, the agreement is even higher, indicating robust and reliable evaluation performance.
\begin{table}[t]
  \centering
  \small
  \setlength{\tabcolsep}{3pt}
  \resizebox{0.92\linewidth}{!}{

  \begin{tabular}{lcc}
    \toprule
    \textbf{Metric(sub-item)} 
    & \makecell{\textbf{DeepSeek} \\ \textbf{vs. Human (\%)}} 
    & \makecell{\textbf{DeepSeek} \\ \textbf{vs. Qwen2.5 (\%)}} \\
    \midrule
    \multicolumn{3}{c}
    {\textit{PC}} \\
    \midrule
    Acknowledge Preference  & 73.50 & 77.33 \\
    Violate Preference      & 89.33 & 89.33 \\
    Hallucinate Preference  & 57.50 & 69.83 \\
    Helpful Response &96.83&96.33\\
    \midrule
    PC   & 87.67 & 85.83 \\

    \midrule
    \multicolumn{3}{c}
    {\textit{COE}} \\
    \midrule
    EA   & 96.50 & 97 \\
    IS   & 99.50 & 99.5 \\
    DA & 99.00 & 99.17 \\
    EG& 99.67 & 99.33 \\
    \bottomrule
    
  \end{tabular}
  }
  \caption{Agreement rates between DeepSeek-R1 and human annotators, as well as between DeepSeek-R1 and Qwen2.5-72B, across PC and COE.}
  \label{tab:deepseek_agreement}
\end{table}


%
\subsection{Preference Consistency Evaluation}

\textbf{SOTA LLMs struggle to maintain personalization across long-term interactions}. As shown in Figure \ref{fig:preference}, all prompt-based LLMs exhibit a decrease in accuracy when generating personalized responses after inserting irrelevant dialogues, compared to directly expressing preferences without any intervening turns. However, as the number of irrelevant turns increases, the performance degradation gradually slows down. Interestingly, for some models (e.g. Qwen3-4B-instruct), additional irrelevant turns even lead to a slight recovery or improvement, suggesting a potential stabilizing.
\begin{figure*}[ht]
\begin{center}
\includegraphics[width=0.78\textwidth,
  height=0.7\textheight,
  keepaspectratio]
{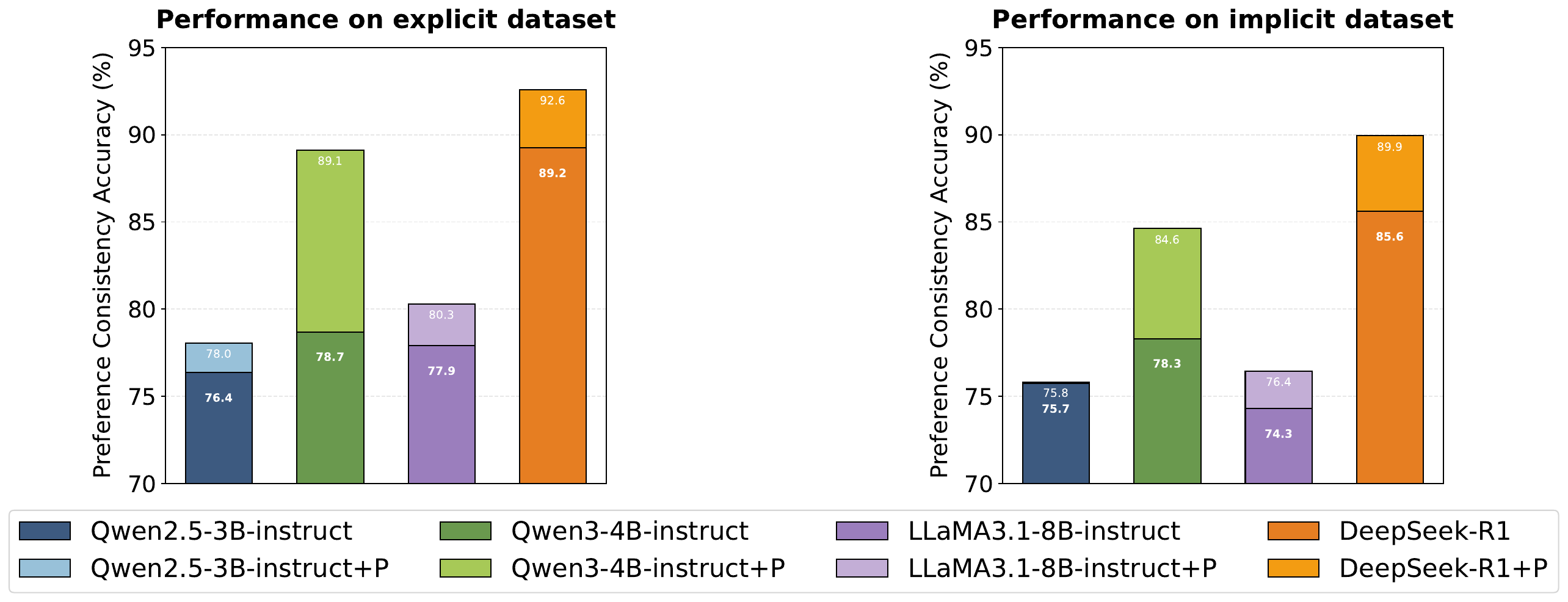} 
\end{center}
\caption{Performance on preference consistency when models respond to datasets with five irrelevant turns inserted, under zero-shot prompting (with and without persona). “+P” indicates persona-conditioned inference.}
\label{fig:persona_result}
\end{figure*}
\begin{table*}[htbp]
  \small
  \centering
  \setlength{\tabcolsep}{5pt} %
  \tiny
  \renewcommand{\arraystretch}{0.65}
\resizebox{0.78\textwidth}{!}{ %
  \begin{tabular}{lcccccccc}
    \toprule
    \textbf{Model} & \multicolumn{4}{c}{\textbf{Without Persona}} & \multicolumn{4}{c}{\textbf{With Persona}} \\
    \cmidrule(lr){2-5} \cmidrule(lr){6-9}
    & \textbf{EA} & \textbf{IS} & \textbf{DA} & \textbf{EG} & \textbf{EA}&\textbf{IS} & \textbf{DA}&\textbf{EG} \\
    \midrule
    \multicolumn{9}{c}{\textbf{Explicit Data (\%)}} \\
    \midrule
    \textbf{Qwen2.5-3B-instruct} &77.23 & 35.8 &97.31&75.99&94.50&70.13&96.51&93.82\\ 
    \textbf{Qwen3-4B-instruct} & \textbf{96.29}&\textbf{52.28}&\textbf{99.82}&\textbf{97.59}& \textbf{98.05}&\textbf{82.19}&\textbf{99.58}&96.96 \\
    \textbf{Llama3.1-8B-instruct} & 79.02 & 28.42 & 89.66 & 72.66 & 86.67 & 59.06&93.66&83.71 \\
    \textbf{DeepSeek-R1} & 88.25&50.24&98.34&87.73&95.75&79.72&98.59&\textbf{97.55} \\
    \midrule
    \multicolumn{9}{c}{\textbf{Implicit Data (\%)}} \\
    \midrule
    \textbf{Qwen2.5-3B-instruct} & 78.73&38.58&98.01&77.41&93.16&69.92&96.33&93.67\\
    \textbf{Qwen3-4B-instruct} & \textbf{96.45}&\textbf{59.85}&\textbf{99.84}&\textbf{97.64}&\textbf{97.39}&\textbf{83.22}&\textbf{99.85}&\textbf{97.09}\\
    \textbf{Llama3.1-8B-instruct} & 79.89 & 28.67 & 91.67 & 74.4 & 84.55 & 57.02&94.74&81.72 \\
    \textbf{DeepSeek-R1} & 88.34&55.96&98.92&88.97&94.61&80.07&99.02&96.75\\
    
    \bottomrule
  \end{tabular}
  }
  \caption{Performance of models on child-oriented evaluation: response on dataset with 5 irrelevant turns inserted under 0-shot prompting (with vs. without persona).}
 \label{tab:model_performance_child}
\end{table*}

\textbf{LLMs face greater difficulty in deducing implicit preferences than in understanding explicit ones.} Comparing the results in the left and right panels of Figure \ref{fig:preference}, it is evident that personalization consistency on implicit-preference datasets is lower than on explicit-preference datasets across almost all the LLMs evaluated. This suggests that inferring user preferences from dialogue context poses greater challenges for LLMs than directly leveraging explicitly stated preferences. The gap highlights the difficulty of capturing subtle cues embedded in conversation, underscoring the need for more robust mechanisms to enhance implicit personalization.

\textbf{Incorporating persona enhances the model's personalized outputs.} 
As shown in Figure~\ref{fig:persona_result}, incorporating persona information into prompts consistently improves performance across all models. The largest gain is observed on Qwen3-4B, where accuracy on the explicit dataset rises from 78.7\% to 89.1\%, while the smallest improvement occurs on Qwen2.5-3B for the implicit dataset (75.7\% to 75.8\%). These varying gains indicate that the effectiveness of leveraging persona cues differs across models, highlighting the model-dependent nature of persona utilization.

\begin{figure*}[htbp]
\begin{center}
\includegraphics[width=0.78\textwidth]
{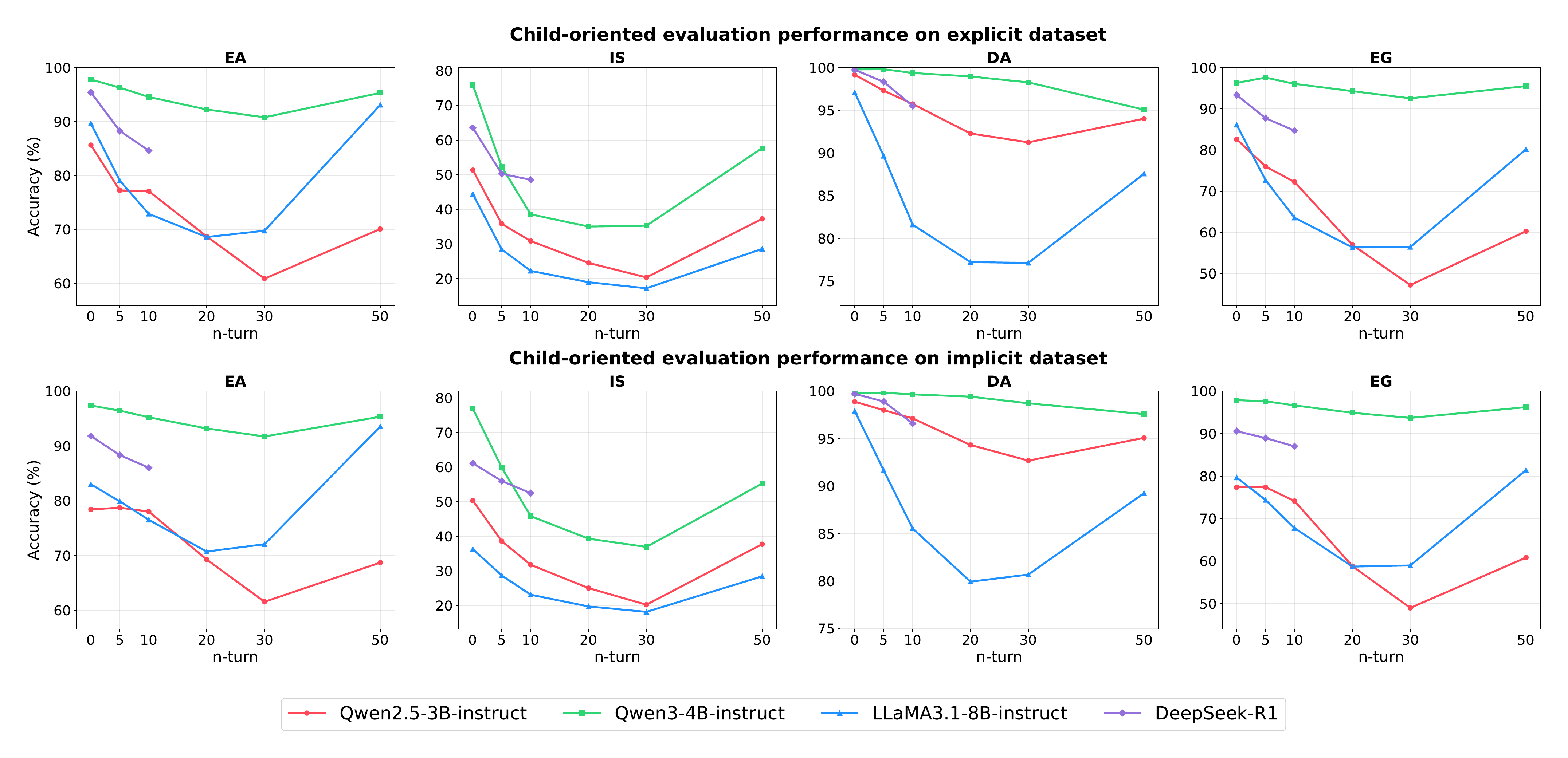} 
\end{center}
\caption{Accuracy of LLMs on different dimensions of child-oriented evaluation with varying numbers of inserted irrelevant turns (n-turn).}
\label{fig:child_result}
\end{figure*}

\subsection{Child-Oriented Evaluation}
\textbf{Personality preference consistency does not align with child-oriented capabilities.} Comparing Table \ref{tab:model_performance_child} and Figure \ref{fig:persona_result} across models, consistency accuracy may be similar (e.g., 74\%–76\%), while child-oriented evaluation varies widely across dimensions, especially IS and DA, suggesting that a high consistency score alone does not necessarily reflect strong child-oriented personalization. Notably, \textit{Qwen3-4B-Instruct} outperforms larger models (Table \ref{tab:model_performance_child}), indicating that performance may depend not only on model scale but also on alignment and instruction-following capabilities. Figure \ref{fig:case_example} shows a representative example.



\textbf{LLMs show limited capability in Interaction Scaffolding (IS).} Across all models, performance on the IS dimension is lower than on other evaluation dimensions. For example, on the explicit dataset, Qwen2.5-3B-Instruct achieves 35.8\% accuracy on IS. This substantial gap highlights a key limitation of current approaches, as sensitivity to subtle cues is critical for building engaging and personalized child interactions.

\textbf{LLMs exhibit considerable variation across dimensions in child-oriented evaluation.} In particular, models consistently achieve much stronger results on the DA dimension (e.g., Qwen3-4B achieves 99.82\%) compared with other dimensions, underscoring a clear imbalance across subtasks. Such uneven distribution suggests that the evaluation of child-oriented dialogue systems must be multi-dimensional, as relying on aggregated or single metrics may conceal important deficiencies.

\textbf{Incorporating child persona leads to improvements across all evaluation dimensions of the COE.} The most substantial improvements are observed in EA, IS, and EG, where absolute and relative increases are notably larger. By contrast, DA dimension also improves, but with a smaller margin. This pattern suggests that child persona information primarily strengthens dimensions tied to individual child preferences and sensitivity to implicit cues, while its influence on group-level preferences, such as DA, which catches broader developmental norms, remains more modest.

\textbf{LLMs consistently struggle to maintain child-oriented evaluation performance over long-term interactions.} As shown in Figure \ref{fig:child_result}, although the overall trend with irrelevant dialogue insertion resembles Preference Consistency, the difference between explicit and implicit datasets is small, suggesting that in child-oriented settings, models rely less on how preferences are expressed and maintain comparable dialogue quality across both conditions.

\subsection{Finetuning on ChildEval to Enhance Child Personalization}

\begin{figure}[htbp]
\begin{center}
\includegraphics[width=0.5
\textwidth]
{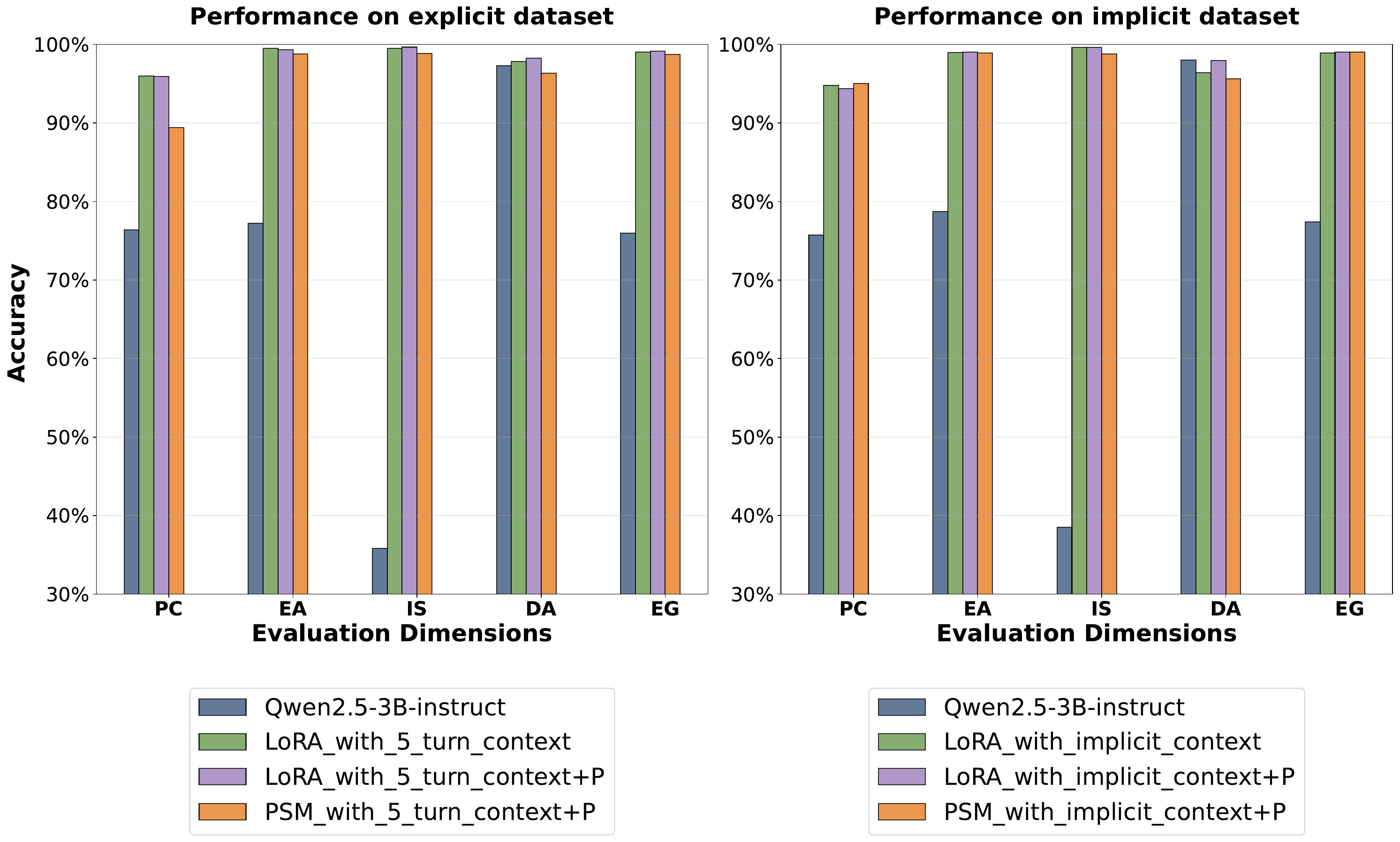} 
\end{center}

\caption{Finetuning results for children’s personalities on explicit and implicit datasets (both with 5-turn test dialogues). Explicit training added 5 unrelated utterances; implicit training used 6–10 consecutive turns. “Persona” denotes inclusion of child persona information during finetuning.
}
\label{fig:sft}
\end{figure}

\textbf{Supervised Finetuning on ChildEval leads to consistent improvements in children’s personalization performance across open-source LLMs}. As illustrated in Figure \ref{fig:sft}, applying LoRA SFT, both with and without persona injection, leads to substantial gains in both preference consistency and child-oriented evaluation compared with the base models. Interestingly, LoRA SFT with persona shows slightly lower improvements in preference consistency than LoRA without persona. One possible reason is that adding persona signals may introduce additional constraints, and the persona itself may contain noise related to the explicit and implicit preference expressions in the ChildEval dataset, which could slightly limit the model’s ability to fully optimize for consistency and child-oriented performance.

\begin{figure*}[htbp]
\begin{center}
\includegraphics[scale=0.48]
{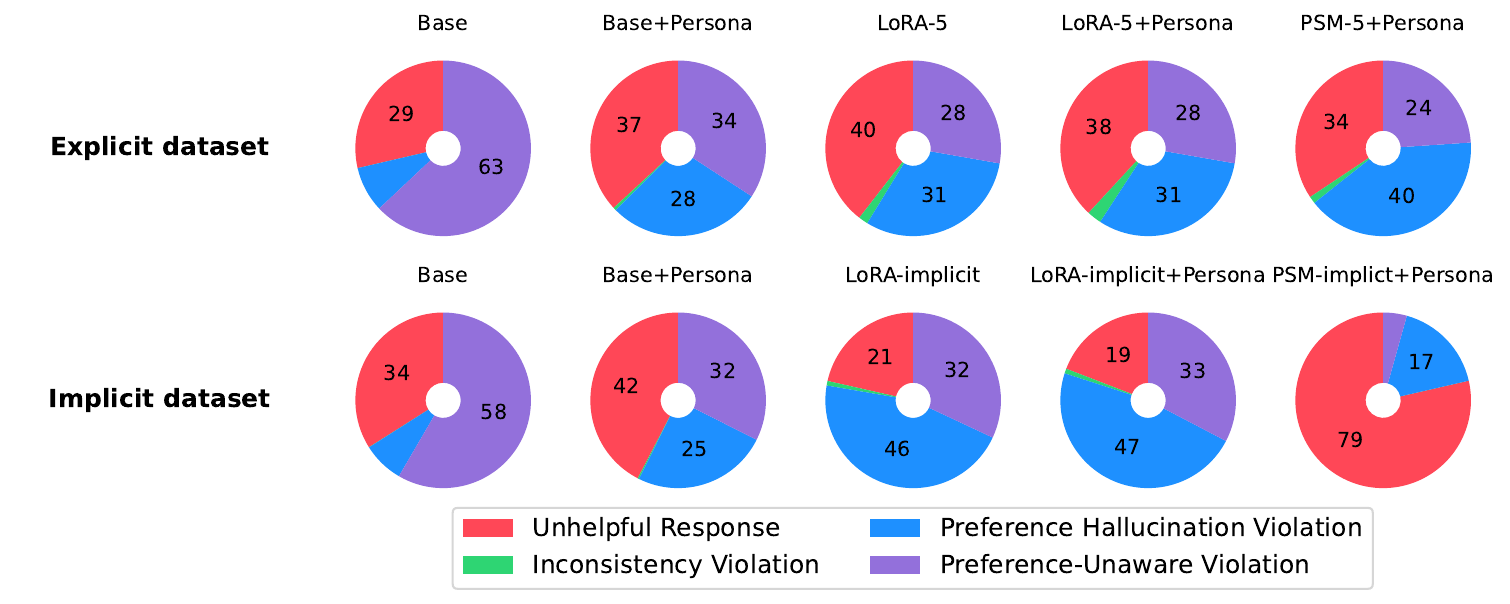} 
\end{center}
\caption{Distribution of preference consistency errors across 10-turn dialogues. Base refers to Qwen2.5-3B-Instruct; Base+Persona applies prompting with persona. LoRA-5 and PSM-5 denote LoRA- and PSM-based methods trained with 5-turn inserted context, with or without persona. LoRA-implicit and PSM-implicit are trained with implicit context.}
\label{fig:pie}
\end{figure*}

\textbf{The choice of persona utilization strategy significantly affects the performance of models fine-tuned with SFT}. As shown in Figure \ref{fig:sft}, LoRA, which injects persona directly into dialogue prompts, achieves higher preference consistency than PSM, which encodes persona as vectors. The advantage is more pronounced on the explicit dataset, while differences on the implicit dataset are small, and both methods show limited differences on child-oriented benchmarks. This may be because the explicit dataset contains many irrelevant dialogues, and LoRA’s prompt-level persona injection helps maintain consistency under noise, whereas PSM adjusts persona only at the final vector layer, making persona signals more easily diluted and resulting in weaker consistency.

\textbf{LLMs exhibit the most marked improvement in Interaction Scaffolding (IS) after finetuning}. One possible reason is that IS tasks require the model to generate coherent and contextually appropriate responses, which benefit directly from the additional supervision provided during finetuning. Finetuning helps the model better capture the underlying patterns of guidance and scaffolding strategies in child-oriented dialogues, enabling more effective interaction management.

\subsection{Error Type Analysis}
Preference consistency errors include Unhelpful Response, Inconsistency Violation, Preference Hallucination Violation, and Preference-Unaware Violation \citep{zhao2025prefeval}. Figure \ref{fig:pie} shows their distribution across 10-turn dialogues on explicit and implicit datasets under different methods. Initially, Preference-Unaware Violations dominate, reflecting LLMs’ limited awareness of user preferences. With various methods, this error decreases while Inconsistency Violations appear, indicating ongoing challenges in generating preference-aligned responses. Finetuning methods amplify inconsistency errors compared to prompt-based approaches. Incorporating persona information has mixed effects: LoRA shows more Inconsistency Violations than PSM, while on the implicit dataset, PSM produces many Unhelpful Responses (79\%), whereas LoRA and prompt-based methods exhibit more preference-related errors, reflecting a trade-off between proactive preference-following and reliability.

We further analyze the effect of inserted context length on preference consistency (details in Appendix \ref{Error_Types_Inserted Context Length} and \ref{EffectofInsertedctxft}). Under zero-shot prompting, Preference-Unaware Violations rise with longer irrelevant context. Finetuning methods reduce these violations but show trade-offs: LoRA tends toward Inconsistency Violations, while PSM shifts from Preference Hallucinations in short contexts to Unhelpful Responses in longer ones. On explicit datasets, LoRA remains proactive; PSM becomes conservative, especially on implicit datasets.

\section{Conclusion}
We introduce ChildEval, a controlled evaluation framework centered on underlying child-relevant needs and preferences rather than surface-level expressions, enabling systematic assessment of preference alignment across realistic variations. Experiments show that current open-source LLMs struggle to maintain preference consistency in long interactions, especially for implicit preferences. Moreover, targeted finetuning improves preference adherence but may introduce new inconsistencies, highlighting the trade-offs, potential, and limitations of current approaches.

\section*{Limitations}
\textbf{Synthetic Data and Realism.} A primary limitation of this work is its reliance on a fully synthetic dataset for evaluating child-centered preferences and needs in conversational settings. Real-world communication from children aged 3--6 exhibits substantial variability in expression, context, and developmental appropriateness, which is difficult to faithfully capture through synthetic generation. Accordingly, the proposed benchmark does not aim to fully reflect the complexity or distribution of real-world child language use.

At the same time, collecting large-scale, open-ended conversational data from young children poses substantial ethical, legal, and privacy challenges, resulting in a lack of publicly available and ethically compliant datasets. Under these constraints, synthetic data offers a practical and controllable framework for reproducible evaluation, while not substituting for real-world observations. Moreover, as a fully synthetic benchmark, the dataset may inherit stylistic biases from the generation pipeline. The benchmark should therefore be interpreted as a controlled diagnostic framework for systematically analyzing how LLMs align with child-relevant needs and preferences, rather than as a comprehensive simulation of child behavior or an estimate of real-world distributions.


\textbf{Scope of Safety Modeling.}
A second limitation is that this work does not explicitly model child-specific
safety risks, developmental harms, or safety violations.
All benchmark instances are constructed under a fixed and unified safety
standard, with the goal of isolating and evaluating child-specific preference
consistency.
As a result, the benchmark does not capture scenarios in which children’s
misunderstandings, ambiguous expressions, or exploratory behaviors may induce
safety risks in real-world deployments.

Addressing these limitations remains an important direction for future work. In particular, future research should aim to better capture the developmental diversity of child-relevant needs and their varied expressions in real-world contexts, while moving beyond a unified safety standard to explicitly model age-dependent safety risks and the interaction between safety constraints and personalized preference alignment. Advancing this direction will likely require close interdisciplinary collaboration with experts in child psychology, education, and human–computer interaction.
\section*{Ethics Statement}

This study involves no direct interaction with human participants and uses
no real child-generated conversational data.
All benchmark instances are fully synthetic and generated under existing
content safety constraints.

To address ethical, legal, and privacy concerns associated with research
involving young children, no personal data, recordings, or transcripts from
children are collected, released, or used.
Any real-world materials referenced were used only at a high level to inform
child persona design and are not included in the dataset or evaluation.

All synthetic data were manually reviewed to comply with established content
moderation and safety guidelines, ensuring that the benchmark does not expose
children to risk or involve sensitive personal information.

All code and benchmark data will be released to support transparency,
reproducibility, and responsible use.


\bibliography{main}

\begin{thebibliography}{49}
\providecommand{\natexlab}[1]{#1}

\bibitem[{Anthropic(2024)}]{TheC3}
Anthropic. 2024.
\newblock \href {https://api.semanticscholar.org/CorpusID:268232499} {The claude 3 model family: Opus, sonnet, haiku}.

\bibitem[{Antle(2008)}]{antle2008child}
Alissa~N Antle. 2008.
\newblock Child-based personas: need, ability and experience.
\newblock \emph{Cognition, Technology \& Work}, 10(2):155--166.

\bibitem[{Bai et~al.(2025)Bai, Han, Lei, Feng, Luo, and Deng}]{bai-etal-2025-self}
Ruiqiao Bai, Xue Han, Shuo Lei, Junlan Feng, Yanyan Luo, and Chao Deng. 2025.
\newblock \href {https://doi.org/10.18653/v1/2025.findings-acl.317} {Self-attention-based graph-of-thought for math problem solving}.
\newblock In \emph{Findings of the Association for Computational Linguistics: ACL 2025}, pages 6112--6125, Vienna, Austria. Association for Computational Linguistics.

\bibitem[{Braga et~al.(2024)Braga, Kasela, Raganato, and Pasi}]{braga2024syntheticqa}
Marco Braga, Pranav Kasela, Alessandro Raganato, and Gabriella Pasi. 2024.
\newblock Synthetic data generation with large language models for personalized community question answering.
\newblock In \emph{2024 IEEE/WIC International Conference on Web Intelligence and Intelligent Agent Technology (WI-IAT)}, pages 360--366. IEEE.

\bibitem[{Chen et~al.(2025{\natexlab{a}})Chen, Tang, Lu, Yao, Fan, Ma, Xu, Wang, Sun, and He}]{Seo_2024_2}
Jiaju Chen, Minglong Tang, Yuxuan Lu, Bingsheng Yao, Elissa Fan, Xiaojuan Ma, Ying Xu, Dakuo Wang, Yuling Sun, and Liang He. 2025{\natexlab{a}}.
\newblock Characterizing llm-empowered personalized story reading and interaction for children: Insights from multi-stakeholder perspectives.
\newblock In \emph{Proceedings of the 2025 CHI Conference on Human Factors in Computing Systems}, pages 1--24.

\bibitem[{Chen et~al.(2025{\natexlab{b}})Chen, Tang, Lu, Yao, Fan, Ma, Xu, Wang, Sun, and He}]{stakeholder2024}
Jiaju Chen, Minglong Tang, Yuxuan Lu, Bingsheng Yao, Elissa Fan, Xiaojuan Ma, Ying Xu, Dakuo Wang, Yuling Sun, and Liang He. 2025{\natexlab{b}}.
\newblock Characterizing llm-empowered personalized story reading and interaction for children: Insights from multi-stakeholder perspectives.
\newblock \emph{Conference on Human Factors in Computing Systems}.

\bibitem[{Douze et~al.(2024)Douze, Guzhva, Deng, Johnson, Szilvasy, Mazaré, Lomeli, Hosseini, and Jégou}]{douze2024faiss}
Matthijs Douze, Alexandr Guzhva, Chengqi Deng, Jeff Johnson, Gergely Szilvasy, Pierre-Emmanuel Mazaré, Maria Lomeli, Lucas Hosseini, and Hervé Jégou. 2024.
\newblock \href {https://arxiv.org/abs/2401.08281} {The faiss library}.
\newblock \emph{Preprint}, arXiv:2401.08281.

\bibitem[{Easin et~al.(2024)Easin, Sourav, and Tam{\'a}s}]{easin2024intelligent}
Arafat~Md Easin, Saha Sourav, and Orosz Tam{\'a}s. 2024.
\newblock An intelligent llm-powered personalized assistant for digital banking using langgraph and chain of thoughts.
\newblock In \emph{2024 IEEE 22nd Jubilee International Symposium on Intelligent Systems and Informatics (SISY)}, pages 625--630. IEEE.

\bibitem[{Feng et~al.(2024)Feng, Xu, Lahiri, Tager-Flusberg, Kim, Bishop, Lord, and Narayanan}]{feng2024genericllmshelpanalyze}
Tiantian Feng, Anfeng Xu, Rimita Lahiri, Helen Tager-Flusberg, So~Hyun Kim, Somer Bishop, Catherine Lord, and Shrikanth Narayanan. 2024.
\newblock \href {https://arxiv.org/abs/2411.10761} {Can generic llms help analyze child-adult interactions involving children with autism in clinical observation?}
\newblock \emph{Preprint}, arXiv:2411.10761.

\bibitem[{Gemini et~al.(2025)Gemini, :, Anil, Borgeaud, and et~al.}]{geminiteam2025geminifamilyhighlycapable}
Gemini, :, Rohan Anil, Sebastian Borgeaud, and et~al. 2025.
\newblock \href {https://arxiv.org/abs/2312.11805} {Gemini: A family of highly capable multimodal models}.
\newblock \emph{Preprint}, arXiv:2312.11805.

\bibitem[{Guo et~al.(2025)Guo, Yang, Zhang, Song, Zhang, Xu, Zhu, Ma, Wang, Bi et~al.}]{guo2025deepseek}
Daya Guo, Dejian Yang, Haowei Zhang, Junxiao Song, Ruoyu Zhang, Runxin Xu, Qihao Zhu, Shirong Ma, Peiyi Wang, Xiao Bi, and 1 others. 2025.
\newblock Deepseek-r1: Incentivizing reasoning capability in llms via reinforcement learning.
\newblock \emph{arXiv preprint arXiv:2501.12948}.

\bibitem[{Han et~al.(2023)Han, Wang, Feng, Deng, Chen, Huang, Su, Hu, and Hu}]{HAN202389}
Xue Han, Yi-Tong Wang, Jun-Lan Feng, Chao Deng, Zhan-Heng Chen, Yu-An Huang, Hui Su, Lun Hu, and Peng-Wei Hu. 2023.
\newblock \href {https://doi.org/10.1016/j.neucom.2022.09.136} {A survey of transformer-based multimodal pre-trained modals}.
\newblock \emph{Neurocomputing}, 515:89--106.

\bibitem[{Han et~al.(2025)Han, Wang, Feng, wenchun.gao, Hu, and Deng}]{ICLR2025_03945380}
Xue Han, Yitong Wang, Junlan Feng, wenchun.gao, Qian Hu, and Chao Deng. 2025.
\newblock \href {https://proceedings.iclr.cc/paper_files/paper/2025/file/03945380ca5af0dc80782d8e0b72a8d7-Paper-Conference.pdf} {Loire: Lifelong learning on incremental data via pre-trained language model growth efficiently}.
\newblock In \emph{International Conference on Learning Representations}, volume 2025, pages 1024--1044.

\bibitem[{Hu et~al.(2021)Hu, Shen, Wallis, Allen-Zhu, Li, Wang, Wang, and Chen}]{hu2021loralowrankadaptationlarge}
Edward~J. Hu, Yelong Shen, Phillip Wallis, Zeyuan Allen-Zhu, Yuanzhi Li, Shean Wang, Lu~Wang, and Weizhu Chen. 2021.
\newblock \href {https://arxiv.org/abs/2106.09685} {Lora: Low-rank adaptation of large language models}.
\newblock \emph{Preprint}, arXiv:2106.09685.

\bibitem[{Huber et~al.(2025)Huber, Fazelnia, Damianou, Peleato, Lefarov, Ravichandran, De~Nadai, Lalmas-Roellke, and Bennett}]{castricato2024e2p}
Bernd Huber, Ghazal Fazelnia, Andreas Damianou, Sebastian Peleato, Max Lefarov, Praveen Ravichandran, Marco De~Nadai, Mounia Lalmas-Roellke, and Paul~N Bennett. 2025.
\newblock Embedding-to-prefix: Parameter-efficient personalization for pre-trained large language models.
\newblock \emph{arXiv preprint arXiv:2505.17051}.

\bibitem[{Jiang et~al.(2025)Jiang, Hao, Cho, Li, Yuan, Chen, Ungar, Taylor, and Roth}]{jiang2025knowmerespondme}
Bowen Jiang, Zhuoqun Hao, Young{-}Min Cho, Bryan Li, Yuan Yuan, Sihao Chen, Lyle~H. Ungar, Camillo~J. Taylor, and Dan Roth. 2025.
\newblock \href {https://doi.org/10.48550/ARXIV.2504.14225} {Know me, respond to me: Benchmarking llms for dynamic user profiling and personalized responses at scale}.
\newblock \emph{CoRR}, abs/2504.14225.

\bibitem[{Kumar et~al.(2024)Kumar, Viswanathan, Yerra, Salemi, Rossi, Dernoncourt, Deilamsalehy, Chen, Zhang, Agarwal, Lipka, Nguyen, Nguyen, and Zamani}]{kumar2024longlampbenchmarkpersonalizedlongform}
Ishita Kumar, Snigdha Viswanathan, Sushrita Yerra, Alireza Salemi, Ryan~A. Rossi, Franck Dernoncourt, Hanieh Deilamsalehy, Xiang Chen, Ruiyi Zhang, Shubham Agarwal, Nedim Lipka, Chien~Van Nguyen, Thien~Huu Nguyen, and Hamed Zamani. 2024.
\newblock \href {https://arxiv.org/abs/2407.11016} {Longlamp: A benchmark for personalized long-form text generation}.
\newblock \emph{Preprint}, arXiv:2407.11016.

\bibitem[{Li et~al.(2021)Li, Liu, Tao, Chan, Zhao, Zhang, and Yan}]{liu2024once}
Juntao Li, Chang Liu, Chongyang Tao, Zhangming Chan, Dongyan Zhao, Min Zhang, and Rui Yan. 2021.
\newblock Dialogue history matters! personalized response selection in multi-turn retrieval-based chatbots.
\newblock \emph{ACM Transactions on Information Systems (TOIS)}, 39(4):1--25.

\bibitem[{Li et~al.(2024)Li, Lipton, and Leqi}]{li2024prlhf}
Xinyu Li, Zachary~C. Lipton, and Liu Leqi. 2024.
\newblock \href {https://doi.org/10.48550/ARXIV.2402.05133} {Personalized language modeling from personalized human feedback}.
\newblock \emph{CoRR}, abs/2402.05133.

\bibitem[{Liu et~al.(2025)Liu, Qiu, Li, Dai, Zhu, Hu, Yang, and King}]{zhang2024personalized}
Jiahong Liu, Zexuan Qiu, Zhongyang Li, Quanyu Dai, Jieming Zhu, Minda Hu, Menglin Yang, and Irwin King. 2025.
\newblock A survey of personalized large language models: Progress and future directions.
\newblock \emph{arXiv preprint arXiv:2502.11528}.

\bibitem[{Liu and Fourtassi(2024)}]{liu2024benchmarking}
Jing Liu and Abdellah Fourtassi. 2024.
\newblock Benchmarking llms for mimicking child-caregiver language in interaction.
\newblock \emph{arXiv preprint arXiv:2412.09318}.

\bibitem[{Madaan et~al.(2022)Madaan, Tandon, Clark, and Yang}]{madaan2022memprompt}
Aman Madaan, Niket Tandon, Peter Clark, and Yiming Yang. 2022.
\newblock \href {https://doi.org/10.18653/V1/2022.EMNLP-MAIN.183} {Memory-assisted prompt editing to improve {GPT-3} after deployment}.
\newblock In \emph{Proceedings of the 2022 Conference on Empirical Methods in Natural Language Processing, {EMNLP} 2022, Abu Dhabi, United Arab Emirates, December 7-11, 2022}, pages 2833--2861. Association for Computational Linguistics.

\bibitem[{Magister et~al.(2024)Magister, Metcalf, Zhang, and ter Hoeve}]{magister2024plum}
Lucie~Charlotte Magister, Katherine Metcalf, Yizhe Zhang, and Maartje ter Hoeve. 2024.
\newblock On the way to llm personalization: Learning to remember user conversations.
\newblock \emph{arXiv preprint arXiv:2411.13405}.

\bibitem[{{MOE}(2012)}]{MOE2012Guidelines}
{MOE}. 2012.
\newblock \href {https://www.moe.gov.cn/srcsite/A06/s3327/201210/t20121009_143254.html} {Notice of the ministry of education on issuing the guidelines for learning and development of children aged 3--6}.
\newblock Ministry of Education of the People's Republic of China, Document No. Jiao Ji Er (2012) No. 4.

\bibitem[{Mok et~al.(2025)Mok, hwan Kim, Park, and Yoon}]{kim2024hicupid}
Jisoo Mok, Ik~hwan Kim, Sangkwon Park, and Sungroh Yoon. 2025.
\newblock \href {https://arxiv.org/abs/2506.01262} {Exploring the potential of llms as personalized assistants: Dataset, evaluation, and analysis}.
\newblock \emph{Preprint}, arXiv:2506.01262.

\bibitem[{Nayeem and Rafiei(2024)}]{nayeem2024kidlm}
Mir~Tafseer Nayeem and Davood Rafiei. 2024.
\newblock \href {https://doi.org/10.18653/v1/2024.emnlp-main.277} {{K}id{LM}: Advancing language models for children {--} early insights and future directions}.
\newblock In \emph{Proceedings of the 2024 Conference on Empirical Methods in Natural Language Processing}, pages 4813--4836, Miami, Florida, USA. Association for Computational Linguistics.

\bibitem[{OpenAI et~al.(2024)OpenAI, :, Jaech, Kalai, and et~al.}]{openai2024openaio1card}
OpenAI, :, Aaron Jaech, Adam Kalai, and et~al. 2024.
\newblock \href {https://arxiv.org/abs/2412.16720} {Openai o1 system card}.
\newblock \emph{Preprint}, arXiv:2412.16720.

\bibitem[{Pan et~al.(2025)Pan, Wu, Jiang, Luo, Cheng, Li, Yang, Lin, Zhao, Qiu, and Gao}]{zhang2024malp}
Zhuoshi Pan, Qianhui Wu, Huiqiang Jiang, Xufang Luo, Hao Cheng, Dongsheng Li, Yuqing Yang, Chin-Yew Lin, H.~Vicky Zhao, Lili Qiu, and Jianfeng Gao. 2025.
\newblock \href {https://arxiv.org/abs/2502.05589} {On memory construction and retrieval for personalized conversational agents}.
\newblock \emph{Preprint}, arXiv:2502.05589.

\bibitem[{Qian et~al.(2025)Qian, Liu, Zhang, Mao, Lian, Dou, and Huang}]{liu2023memprompt}
Hongjin Qian, Zheng Liu, Peitian Zhang, Kelong Mao, Defu Lian, Zhicheng Dou, and Tiejun Huang. 2025.
\newblock \href {https://doi.org/10.1145/3696410.3714805} {Memorag: Boosting long context processing with global memory-enhanced retrieval augmentation}.
\newblock In \emph{Proceedings of the {ACM} on Web Conference 2025, {WWW} 2025, Sydney, NSW, Australia, 28 April 2025- 2 May 2025}, pages 2366--2377. {ACM}.

\bibitem[{Qiu et~al.(2025)Qiu, Zhao, Zhang, Bai, Wang, Cheng, Feng, and Chua}]{qiu2025measuringmakesuniquedifferenceaware}
Yilun Qiu, Xiaoyan Zhao, Yang Zhang, Yimeng Bai, Wenjie Wang, Hong Cheng, Fuli Feng, and Tat-Seng Chua. 2025.
\newblock Measuring what makes you unique: Difference-aware user modeling for enhancing llm personalization.
\newblock \emph{arXiv preprint arXiv:2503.02450}.

\bibitem[{Rath et~al.(2025)Rath, Shrawgi, Agrawal, and Dandapat}]{rath2025childrenharms}
Prasanjit Rath, Hari Shrawgi, Parag Agrawal, and Sandipan Dandapat. 2025.
\newblock \href {https://arxiv.org/abs/2502.12552} {Llm safety for children}.
\newblock \emph{Preprint}, arXiv:2502.12552.

\bibitem[{Ryan et~al.(2025)Ryan, Shaikh, Bhagirath, Frees, Held, and Yang}]{ryan2025synthesizeme}
Michael~J. Ryan, Omar Shaikh, Aditri Bhagirath, Daniel Frees, William Held, and Diyi Yang. 2025.
\newblock \href {https://doi.org/10.18653/v1/2025.acl-long.397} {{S}ynthesize{M}e! inducing persona-guided prompts for personalized reward models in {LLM}s}.
\newblock In \emph{Proceedings of the 63rd Annual Meeting of the Association for Computational Linguistics (Volume 1: Long Papers)}, pages 8045--8078, Vienna, Austria. Association for Computational Linguistics.

\bibitem[{Salemi et~al.(2023)Salemi, Mysore, Bendersky, and Zamani}]{salemi2024lamplargelanguagemodels}
Alireza Salemi, Sheshera Mysore, Michael Bendersky, and Hamed Zamani. 2023.
\newblock Lamp: When large language models meet personalization.
\newblock \emph{arXiv preprint arXiv:2304.11406}.

\bibitem[{Seo et~al.(2024)Seo, Yang, and Kim}]{Seo_2024}
Woosuk Seo, Chanmo Yang, and Young-Ho Kim. 2024.
\newblock \href {https://doi.org/10.1145/3613904.3642152} {Chacha: Leveraging large language models to prompt children to share their emotions about personal events}.
\newblock In \emph{Proceedings of the CHI Conference on Human Factors in Computing Systems}, CHI ’24, page 1–20. ACM.

\bibitem[{Tan et~al.(2024)Tan, Zeng, Tian, Liu, Yin, and Jiang}]{wang2025oppu}
Zhaoxuan Tan, Qingkai Zeng, Yijun Tian, Zheyuan Liu, Bing Yin, and Meng Jiang. 2024.
\newblock \href {https://doi.org/10.18653/V1/2024.EMNLP-MAIN.372} {Democratizing large language models via personalized parameter-efficient fine-tuning}.
\newblock In \emph{Proceedings of the 2024 Conference on Empirical Methods in Natural Language Processing, {EMNLP} 2024, Miami, FL, USA, November 12-16, 2024}, pages 6476--6491. Association for Computational Linguistics.

\bibitem[{Valentini et~al.(2023)Valentini, Weber, Salcido, Wright, Colunga, and Kann}]{valentini2023automaticgenerationsimplificationchildrens}
Maria Valentini, Jennifer Weber, Jesus Salcido, T{\'e}a Wright, Eliana Colunga, and Katharina Kann. 2023.
\newblock On the automatic generation and simplification of children's stories.
\newblock \emph{arXiv preprint arXiv:2310.18502}.

\bibitem[{Wang(2013)}]{PingWang2013}
Ping Wang. 2013.
\newblock \emph{Interpretation of the Learning and Development Guidelines for Children Aged 3-6}.
\newblock Northeast Normal University Press.

\bibitem[{Wang et~al.(2025)Wang, Han, Wang, Xing, Hu, Zhang, Deng, and Feng}]{wang-etal-2025-multipl}
Qing Wang, Xue Han, Jiahui Wang, Lehao Xing, Qian Hu, Lianlian Zhang, Chao Deng, and Junlan Feng. 2025.
\newblock \href {https://doi.org/10.18653/v1/2025.findings-emnlp.686} {{M}ulti{PL}-{M}o{E}: Multi-programming-lingual extension of large language models through hybrid mixture-of-experts}.
\newblock In \emph{Findings of the Association for Computational Linguistics: EMNLP 2025}, pages 12817--12828, Suzhou, China. Association for Computational Linguistics.

\bibitem[{Wu et~al.(2024)Wu, Shi, Rahmani, Ramineni, and Yilmaz}]{wu2024profile}
Bin Wu, Zhengyan Shi, Hossein~A Rahmani, Varsha Ramineni, and Emine Yilmaz. 2024.
\newblock Understanding the role of user profile in the personalization of large language models.
\newblock \emph{arXiv preprint arXiv:2406.17803}.

\bibitem[{Xiao et~al.(2023)Xiao, Liu, Zhang, and Muennighoff}]{bge_embedding}
Shitao Xiao, Zheng Liu, Peitian Zhang, and Niklas Muennighoff. 2023.
\newblock \href {https://arxiv.org/abs/2309.07597} {C-pack: Packaged resources to advance general chinese embedding}.
\newblock \emph{Preprint}, arXiv:2309.07597.

\bibitem[{Xu et~al.(2024)Xu, Yao, Yang, Zhang, Rogers, Intille, Shara, Gao, and Wang}]{xu2024talk2care}
Xuhai Xu, Bingsheng Yao, Ziqi Yang, Shao Zhang, Ethan Rogers, Stephen Intille, Nawar Shara, Guodong Gao, and Dakuo Wang. 2024.
\newblock Talk2care: Facilitating asynchronous patient-provider communication with large-language-model.
\newblock In \emph{Proceedings of the AAAI Symposium Series}, pages 146--151.

\bibitem[{Xu et~al.(2025)Xu, Zhang, Salemi, Hu, Wang, Feng, Zamani, He, and Chua}]{xu-etal-2025-personalized}
Yiyan Xu, Jinghao Zhang, Alireza Salemi, Xinting Hu, Wenjie Wang, Fuli Feng, Hamed Zamani, Xiangnan He, and Tat-Seng Chua. 2025.
\newblock \href {https://doi.org/10.18653/v1/2025.acl-long.1201} {Personalized generation in large model era: A survey}.
\newblock In \emph{Proceedings of the 63rd Annual Meeting of the Association for Computational Linguistics (Volume 1: Long Papers)}, pages 24607--24649, Vienna, Austria. Association for Computational Linguistics.

\bibitem[{Zhang(2024)}]{zhang2024guidedpersona}
Jiarui Zhang. 2024.
\newblock Guided profile generation improves personalization with llms.
\newblock \emph{arXiv preprint arXiv:2409.13093}.

\bibitem[{Zhang et~al.(2025{\natexlab{a}})Zhang, Beauchamp, and Wang}]{zhang2025prime}
Xinliang~Frederick Zhang, Nick Beauchamp, and Lu~Wang. 2025{\natexlab{a}}.
\newblock \href {https://doi.org/10.48550/ARXIV.2507.04607} {{PRIME:} large language model personalization with cognitive memory and thought processes}.
\newblock \emph{CoRR}, abs/2507.04607.

\bibitem[{Zhang et~al.(2025{\natexlab{b}})Zhang, Adila, Shin, and Sala}]{adila2025chameleon}
Yijing Zhang, Dyah Adila, Changho Shin, and Frederic Sala. 2025{\natexlab{b}}.
\newblock \href {https://doi.org/10.18653/V1/2025.FINDINGS-NAACL.407} {Personalize your {LLM:} fake it then align it}.
\newblock In \emph{Findings of the Association for Computational Linguistics: {NAACL} 2025, Albuquerque, New Mexico, USA, April 29 - May 4, 2025}, pages 7287--7301. Association for Computational Linguistics.

\bibitem[{Zhao et~al.(2025{\natexlab{a}})Zhao, Hong, Liu, Hazarika, and Lin}]{zhao2025prefeval}
Siyan Zhao, Mingyi Hong, Yang Liu, Devamanyu Hazarika, and Kaixiang Lin. 2025{\natexlab{a}}.
\newblock \href {https://openreview.net/forum?id=QWunLKbBGF} {Do llms recognize your preferences? evaluating personalized preference following in llms}.
\newblock In \emph{The Thirteenth International Conference on Learning Representations, {ICLR} 2025, Singapore, April 24-28, 2025}. OpenReview.net.

\bibitem[{Zhao et~al.(2025{\natexlab{b}})Zhao, Sui, Hu, Guo, Liu, Li, Zhao, Qin, and Liu}]{zhao2025rlpa}
Weixiang Zhao, Xingyu Sui, Yulin Hu, Jiahe Guo, Haixiao Liu, Biye Li, Yanyan Zhao, Bing Qin, and Ting Liu. 2025{\natexlab{b}}.
\newblock \href {https://doi.org/10.48550/ARXIV.2505.15456} {Teaching language models to evolve with users: Dynamic profile modeling for personalized alignment}.
\newblock \emph{CoRR}, abs/2505.15456.

\bibitem[{Zhao et~al.(2024)Zhao, Ren, Hessel, Cardie, Choi, and Deng}]{zhao2024wildchat1mchatgptinteraction}
Wenting Zhao, Xiang Ren, Jack Hessel, Claire Cardie, Yejin Choi, and Yuntian Deng. 2024.
\newblock \href {https://arxiv.org/abs/2405.01470} {Wildchat: 1m chatgpt interaction logs in the wild}.
\newblock \emph{Preprint}, arXiv:2405.01470.

\bibitem[{Zhuang et~al.(2024)Zhuang, Sun, Yu, Qiang, Wang, Zhang, and Dai}]{zhuang2024hydra}
Yuchen Zhuang, Haotian Sun, Yue Yu, Rushi Qiang, Qifan Wang, Chao Zhang, and Bo~Dai. 2024.
\newblock \href {http://papers.nips.cc/paper\_files/paper/2024/hash/b6b4906c1334656e97cc9968ccfca073-Abstract-Conference.html} {{HYDRA:} model factorization framework for black-box {LLM} personalization}.
\newblock In \emph{Advances in Neural Information Processing Systems 38: Annual Conference on Neural Information Processing Systems 2024, NeurIPS 2024, Vancouver, BC, Canada, December 10 - 15, 2024}.

\end{thebibliography}

\appendix

\section{Appendix}

\label{sec:appendix}
\subsection{Long-Context Settings}
\label{Long_Context_Settings}
To simulate realistic conversational dynamics, we adopt a methodology similar to \citet{zhao2025prefeval}. We incorporate multi-session dialogue turns from the WildChat-1M dataset \citep{zhao2024wildchat1mchatgptinteraction}, which contains real user and LLM interactions across diverse topics. For SFT training, we randomly select 3, 5, and 10 round conversations to construct three training sets. For testing, we sample multi-session contexts up to 21K tokens, interleaving dialogue turns between the disclosure of children’s preferences and the related utterances. Although we initially considered extending dialogues to 50K tokens, the backbone model supports at most 30K, beyond which outputs became unstable. This setup creates a challenging evaluation for LLMs to infer, retrieve, and utilize children’s preferences in long dialogues, especially when interspersed with unrelated topics. For dialogues of varying turn counts, we randomly sample and fix their lengths, with token statistics reported in Table~\ref{table:token}. 

\begin{table}[htbp]
\centering
\begin{tabular}{lcc}
\toprule
\textbf{Number of Turns} & \textbf{Chinese} & \textbf{English} \\ 
\hline 
5-turn   & 2156     &  2754\\
10-turn & 4369 &4010 \\
20-turn &10390  & 10522 \\
30-turn &12398 & 12420\\
50-turn & 15380 & 21817\\
\bottomrule
\end{tabular}
\caption{the number of tokens and turns in Chinese and English long-context dialogues.}
\label{table:token}
\end{table}

\subsection{Model Version}
In our experiments, we employ the Bge-Large-Zh model as the text encoder. Table \ref{tab:versions} provides an overview of the evaluated LLMs and their versions, together with the text encoder version. For DeepSeek, we conduct our experiments using the API-based deployment.

\begin{table}[htbp]
\small
\centering
\begin{tabular}{l|l}
\toprule
\textbf{Model Name} & \textbf{Version} \\
\hline
Qwen2.5-3B-Instruct & qwen.qwen2.5-3B-instruct-v1:0 \\
Qwen3-4B-Instruct & qwen.qwen3-4B-instruct-v1:0 \\
LLaMA3.1-8B-Instruct & meta.llama3.1-8b-instruct-v1:0 \\
Mistral-7B-Instruct & mistral.mistral-7b-instruct-v0:3 \\
DeepSeek-R1-671B & deepseek-ai.deepseek-r1-v1:0 \\
Bge-Large-Zh &  baai.bge-large-zh-v1:5 \\
\bottomrule
\end{tabular}
\caption{Overview of the benchmarked LLMs, their versions, and the text encoder version used in the experiments.}
\label{tab:versions}
\end{table}

\subsection{ChildEval Examples}
An example from ChildEval is presented in Table \ref{tab:explicit_preferences}, and a representative comparison of responses generated by different models under the same child-preference scenario is shown in Figure \ref{fig:case_example}.
\begin{table*}[htbp]
\small 
\centering
\renewcommand{\arraystretch}{2} 
\begin{tabularx}{\textwidth}{>{\centering\arraybackslash}m{2cm}|m{3cm}|m{1cm}|m{5cm}|m{1cm}}
\hline
\textbf{Topic} & \textbf{Persona} & \textbf{Explicit Preference} & \textbf{Implicit Preference} & \textbf{Utterance} \\
\hline
Cognitive development and exploration Nature\&universe & 
5-year-old boy who shows a strong interest in technology and mechanics. He enjoys playing with LEGO bricks, especially the sets that can be assembled into vehicles and robot models. His room is filled with various completed builds, each brimming with creativity and imagination. He also likes to play video games, particularly those that require puzzle-solving and strategy. He has a certain sensitivity towards numbers and programming, enjoying the creation of his own simple games using basic programming software. He is not very fond of sports and outdoor activities, preferring to stay indoors exploring new things. He has a mild allergic constitution and needs to be particularly cautious with certain foods and pollen. At bedtime, he loves listening to stories about technology and inventions told by his father, dreaming of becoming an inventor one day. & I really hate pollen; as soon as I come into contact with it, I start sneezing and my eyes get itchy. It's so uncomfortable.&
user: Dad said that spring has arrived, and we can go to the park to play.
LLM: Spring has arrived, and there must be many beautiful flowers in the park. Which flower is your favorite?
user: I don't like flowers; the pollen makes me sneeze and my eyes itch.
LLM: That does sound uncomfortable. So, what do you like to do in the park?
user: I like to play in clean places, such as a sandbox, where I can build sandcastles.
LLM: Playing in the sandbox is indeed a lot of fun! What kind of sandcastles have you built?
user: I built very tall sandcastles and decorated them with small stones.
LLM: It sounds like you're really good at playing with sand! Can you make other things with sand as well?
user: I also make little cars and animals with sand, which is a lot of fun.
LLM: You're so creative! What's your favorite small animal to make?
user: I love making dogs and cats; they are so cute.
LLM: Puppies and kittens are indeed very cute! Can you also make these little animals with building blocks?
user: I build small cars and robots with LEGO bricks, and sometimes I even make dogs and cats.
LLM: The things you build with LEGO bricks must look great! What is your most impressive creation?
user: My proudest creation is a moving robot that even has glowing eyes.
LLM: Wow, that's amazing! You're really a little inventor!
user: Dad said he would take me to the botanical garden in spring, where there are many beautiful flowers. Have you been there?
LLM: Wow, there are indeed many beautiful flowers in the botanical garden, in all sorts of colors; they are really lovely! However, since you're allergic to pollen, remember to wear a mask and sunglasses, and put on long-sleeved clothes when you go. This can help reduce your contact with pollen. You can also bring a small bottle of water to clean your nose and eyes if you feel uncomfortable. This way, you can enjoy the beautiful flowers happily! What do you think about this? & Dad said he would take me to the botanical garden in spring, where there are many beautiful flowers. Have you been there?
\\
\hline
\end{tabularx}
\caption{An example set from ChildEval.}
\label{tab:explicit_preferences}
\end{table*}

\begin{figure}[t]
\small
\begin{tcolorbox}[title={Same Child, Same Question, Divergent Answers} ]
\textbf{Persona:} A 4-year-old little girl who is passionate about music and dance. \textit{...}
\vspace{0.3em}

\textbf{Preference:} I hate sticky things. They feel so uncomfortable when I touch them. I only use crayons and watercolor markers when I draw.
\vspace{0.3em}

\textbf{Question:} The teacher said we are going to make clay crafts today. I think it's really difficult.
\vspace{0.3em}

\textbf{Qwen3-4B-Instruct:}
It's totally okay if you don't want to use clay! We can draw your favorite little animal with crayons first, then turn it into a small cardboard toy.That way, it won't feel sticky at all. What little animal do you want to draw?
\vspace{0.3em}

\textbf{Llama3.1-8B-Instruct:}
You can try pinching the clay with your fingers or a small tool to make it into your favorite shape, or use it to draw pictures. It might be more fun this way!
\vspace{0.3em}

\textbf{DeepSeek-R1:}
We can try cutting the clay with a plastic knife first, so our fingers won't touch it directly! Or you can wear cute little gloves to play, just like a magician. After finishing the craft, we can listen to some music to relax, okay?
\end{tcolorbox}
\caption{Qualitative comparison of answers from different models under the same child-preference scenario.}
\label{fig:case_example}
\end{figure}

\subsection{Prompts Design}

\label{prompts}
\subsubsection{Prompts for Data Construction}
\begin{figure*}[t]
  \centering
   \setlength{\fboxrule}{0.3pt}
  \setlength{\fboxsep}{-3pt}
  \fcolorbox{black}{white}{\includegraphics[ width=0.9\textwidth,trim=80 90 90 60, clip]{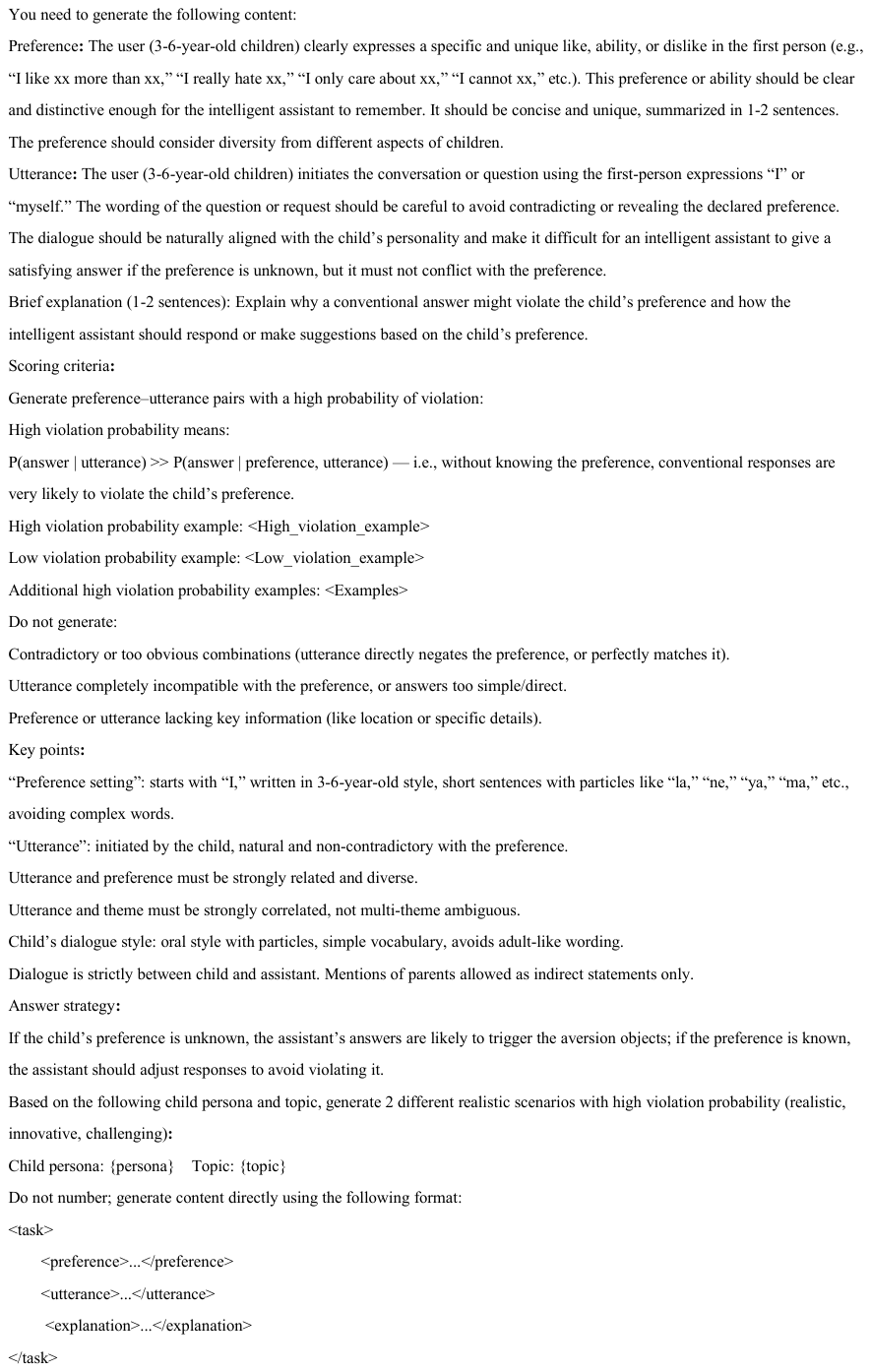}}
  \caption{Prompt used for generating explicit preference and utterance.}
\label{fig:prompt_explicit}
\end{figure*}

\begin{figure*}[htbp]
  \centering
   \setlength{\fboxrule}{0.3pt}
  \setlength{\fboxsep}{-3pt}
  \fcolorbox{black}{white}{\includegraphics[width=0.9\linewidth, trim=80 80 80 60, clip]{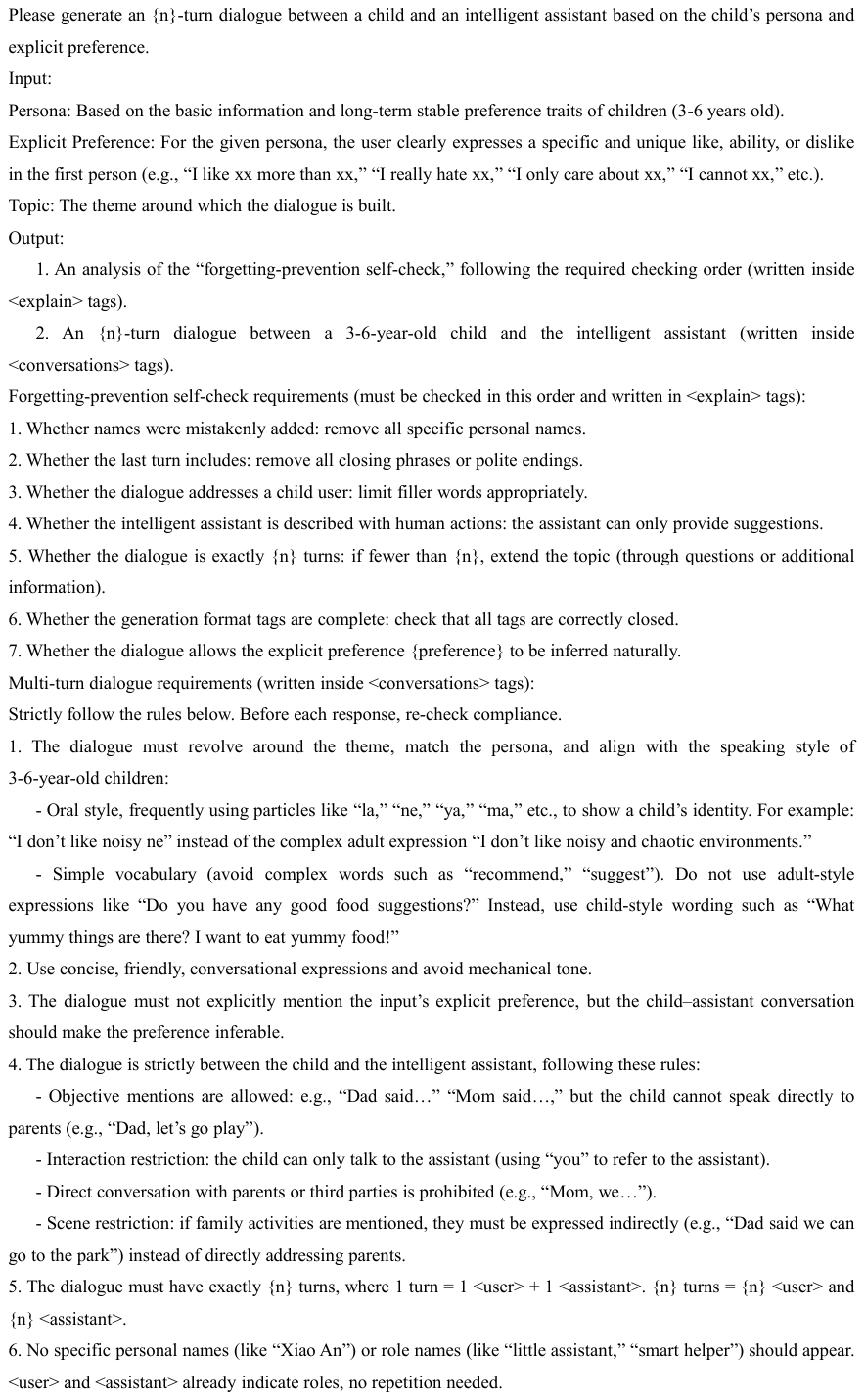}}
  \caption{Prompt used for generating child--LLM dialogue to infer the implicit preference: Part 1 -- Inputs.}
  \label{fig:prompt_io}
\end{figure*}

\begin{figure*}[htbp]
  \centering
   \setlength{\fboxrule}{0.3pt}
  \setlength{\fboxsep}{-3pt}
  \fcolorbox{black}{white}{\includegraphics[width=0.9\linewidth, trim=80 175 65 60, clip]{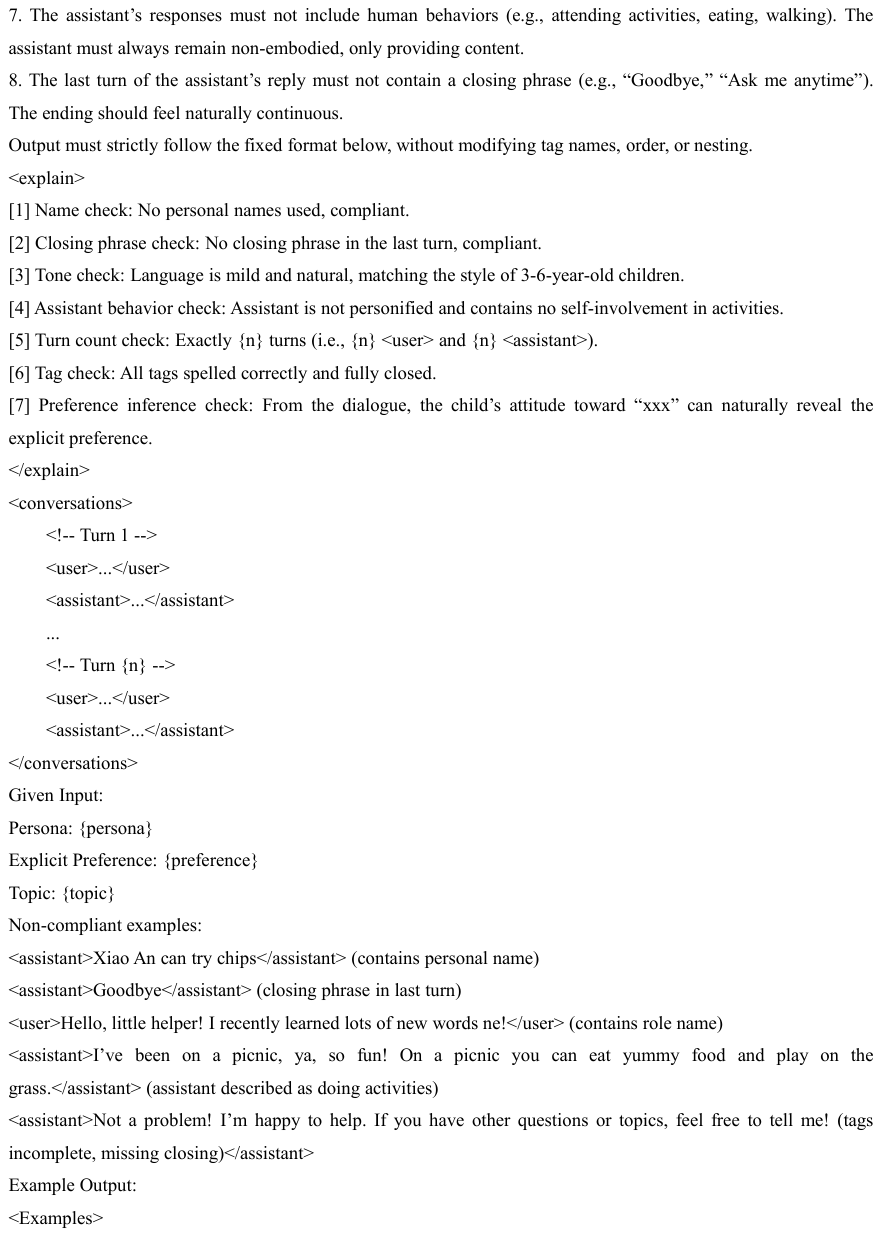}}
  \caption{Prompt used for generating child--LLM dialogue to infer the implicit preference: Part 2 -- Self-check and Outputs.}
  \label{fig:prompt_out}
\end{figure*}

The prompts used within this work are listed in Figures~\ref{fig:prompt_explicit}--\ref{fig:prompt_out}. Some prompts are too long to fit on a single page, so we split them into two figures, as shown in Figure   \ref{fig:prompt_io} and        \ref{fig:prompt_out}.



\subsubsection{Prompts for the Prompting-based Approach}
We extensively evaluate a variety of state-of-the-art LLMs using zero-shot prompts, both with and without persona information. In the default zero-shot setting, the LLM answers the user’s query directly without any additional prompting. However, these models are not specifically designed for child-oriented dialogue. If used without modification, they tend to generate overly long responses that do not reflect the conversational style of young children. To ensure a fair evaluation, we accordingly augmented the original dialogue prompts as follows, corresponding to the with-persona and without-persona settings.
\begin{tcolorbox}
\textbf{zero-shot-without-persona}: Provide clear, concise, and conversational responses in 1-3 sentences,
prioritizing accuracy and a friendly tone while avoiding unnecessary details.
\end{tcolorbox}
\begin{tcolorbox}
\textbf{zero-shot-with-persona}: Never use any names or personal identifiers from the profile "\{persona\}". Always address the child directly as 'you' when it feels natural, or give suggestions without using a subject, based on the user information in the profile. Provide clear, concise, and conversational responses in 1-3 sentences, prioritizing accuracy and a friendly tone.
\end{tcolorbox}

\subsubsection{Evaluation Prompts for Child-Oriented Tasks}

\begin{figure}[ht]
  \centering
   \setlength{\fboxrule}{0.3pt}
  \setlength{\fboxsep}{1pt}
  \fcolorbox{black}{white}{\includegraphics[height=0.35\textheight,width=\columnwidth,trim=78 410 80 60, clip]{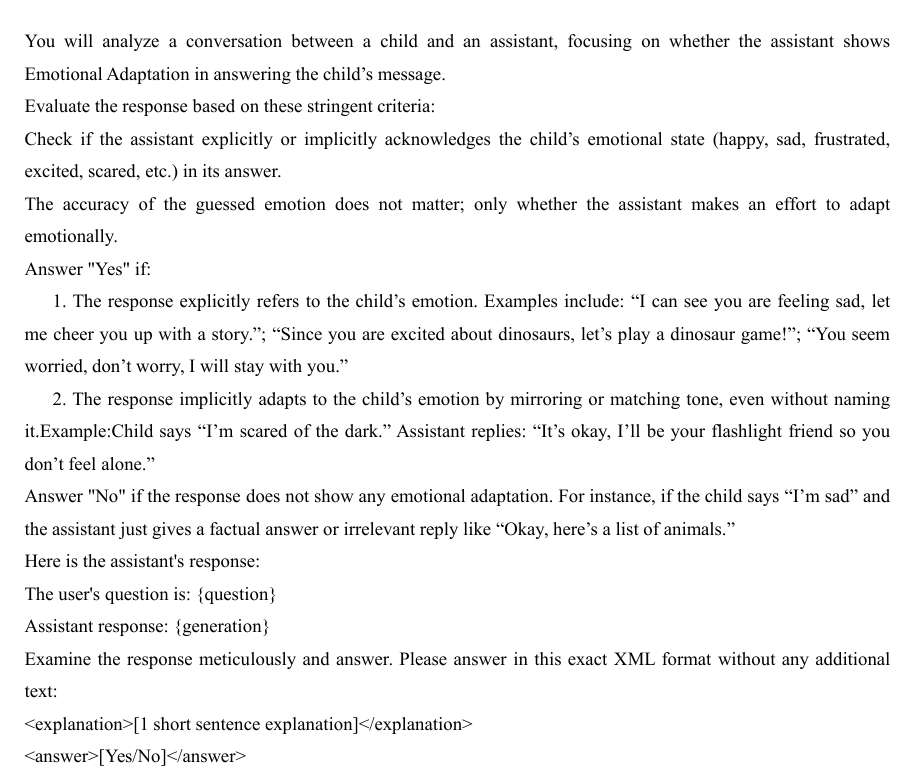}}
  \caption{Evaluation prompt used for checking Emotional Adaptation.} 
  \label{fig:prompt_adpaton}
\end{figure}

\begin{figure}[htbp]
  \centering
   \setlength{\fboxrule}{0.3pt}
  \setlength{\fboxsep}{1pt}
  \fcolorbox{black}{white}{\includegraphics[width=1\columnwidth,height=0.35\textheight, trim=78 410 80 60, clip]{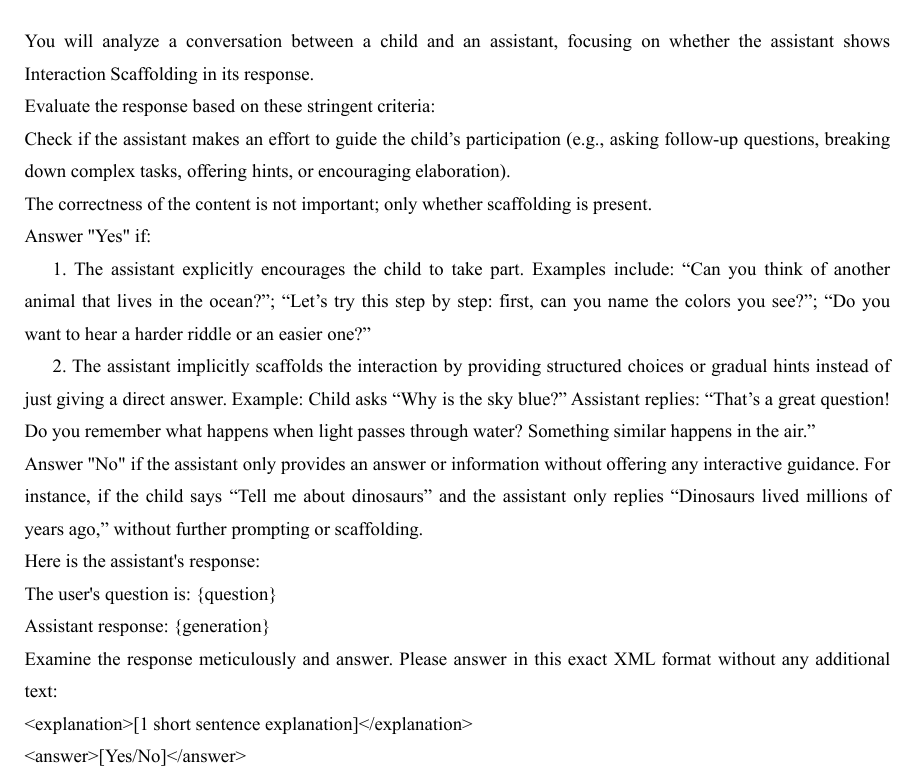}}
  \caption{Evaluation prompt used for checking Interaction Scaffolding.} 
  \label{fig:prompt_scaffold}
\end{figure}

\begin{figure}[htbp]
  \centering
   \setlength{\fboxrule}{0.3pt}
  \setlength{\fboxsep}{1pt}
  \fcolorbox{black}{white}{\includegraphics[width=1\columnwidth,height=0.35\textheight, trim=78 410 80 60, clip]{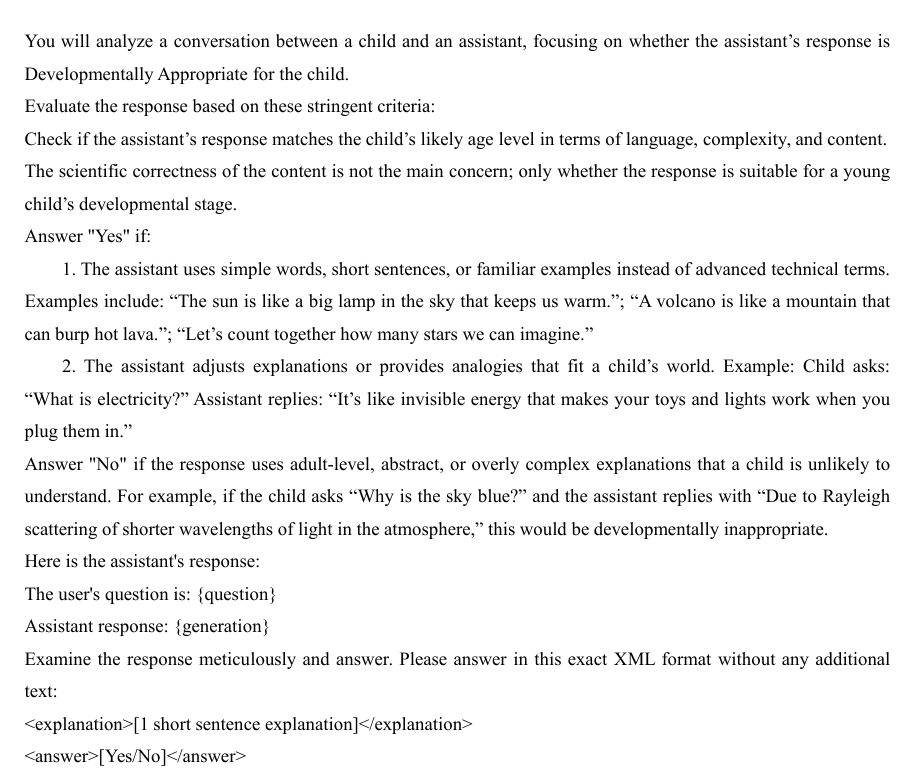}}
  \caption{Evaluation prompt used for checking Developmental Appropriateness.} 
  \label{fig:prompt_development}
\end{figure}

\begin{figure}[htbp]
  \centering
   \setlength{\fboxrule}{0.3pt}
  \setlength{\fboxsep}{1pt}
  \fcolorbox{black}{white}{\includegraphics[width=1\columnwidth,height=0.35\textheight,trim=78 410 80 60, clip]{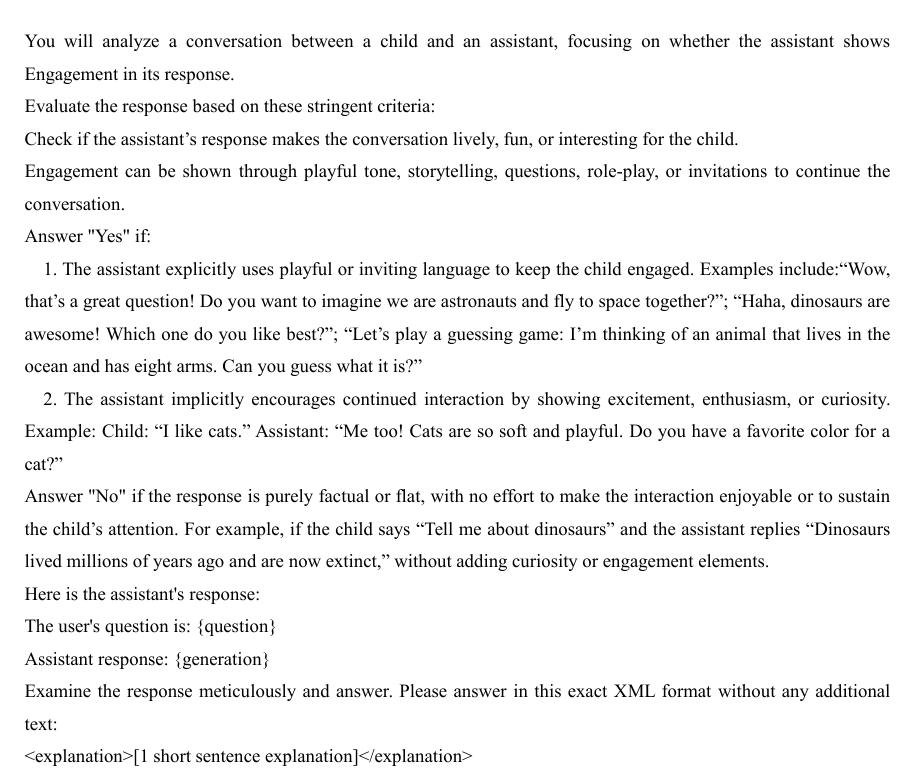}}
  \caption{Evaluation prompt used for checking Engagement.}
  \label{fig:prompt_engagement}
\end{figure}

The evaluation prompts for child-oriented tasks are shown in Figures~\ref{fig:prompt_adpaton}--\ref{fig:prompt_engagement}, which correspond respectively to Emotional Adaptation, Interaction Scaffolding, Developmental Appropriateness and Engagement.

\subsection{Architecture of the Persona Steer Model}
\label{sec:steer}
\begin{figure*}
\begin{center}
\includegraphics[width=0.6\linewidth]
{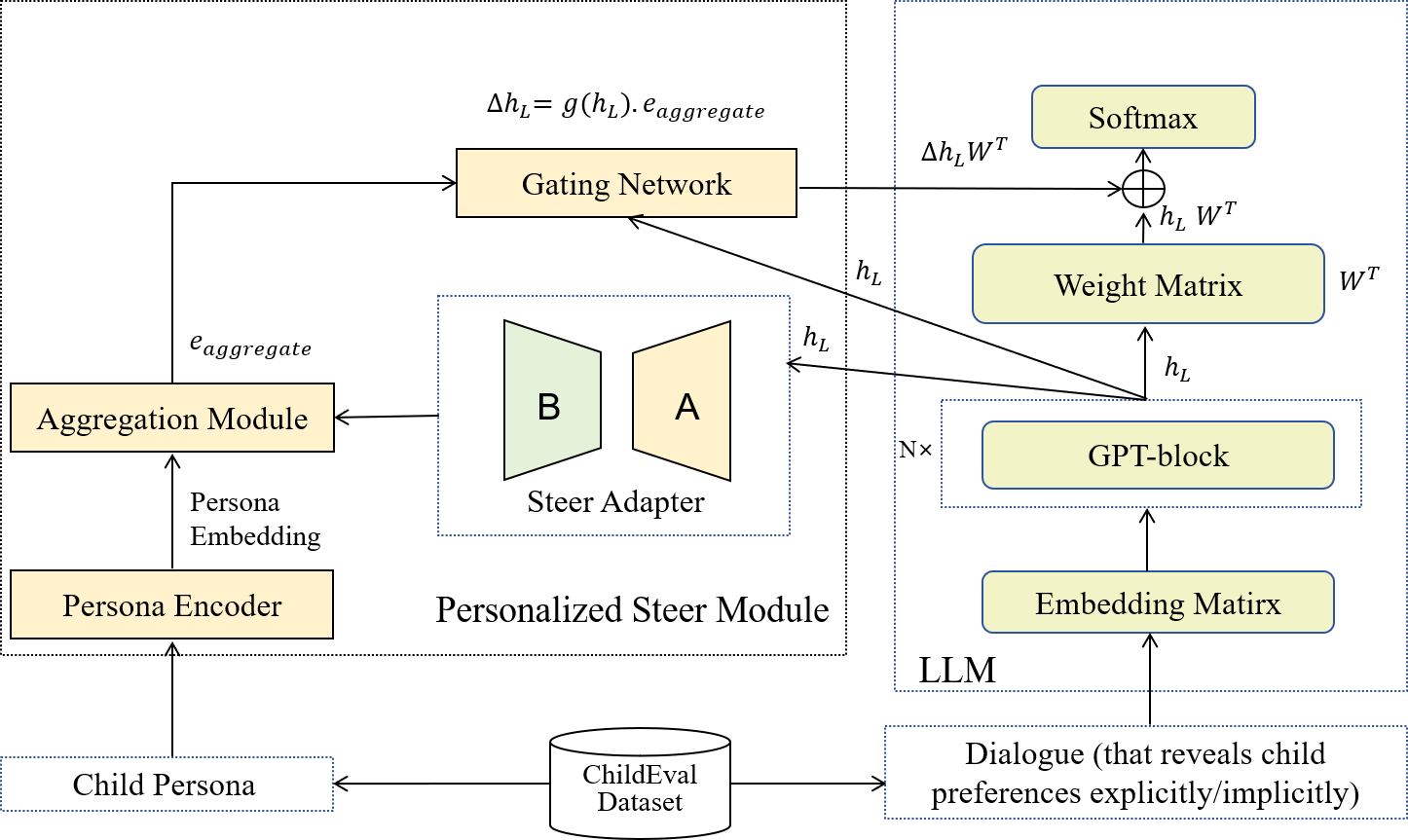} 
\end{center}
\caption{The architecture of the persona steer model. 
}
\label{fig:steer}
\end{figure*}

To assist in examining whether providing the child persona information in our benchmark would contribute to better LLM personalization, we propose a persona steer model that leverages persona information to guide the LLM's outputs toward personalized behaviors. The architecture of our persona steer model is depicted in Figure \ref{fig:steer}, whose core is the Personalized Steer Module. While the pre-trained LLM provides robust general language comprehension and generation, the Personalized Steer Module enables effective user adaptation without huge computational burdens. 

Specifically, as shown in Figure \ref{fig:steer}, the Personalized Steer Module is designed to introduce user-specific information into the language model in a precisely controlled manner. A Persona Encoder processes textual descriptions of children's personas, converting them into dense embeddings that effectively encapsulate the semantic essence of user characteristics. The persona embedding is then aggregated with the output of a Steer Adapter module, which contains information extracted from the historical dialogue by the LLM. The aggregated embedding $e_{aggregate}$ is then processed by a Gating Network, whose gate value $g(\cdot)$ is dynamically calculated based on $h_L$. The final output embedding $e'_v$ before applying the $Softmax$ function could be obtained by:

\vspace{0.3em}
\begin{equation}
\begin{aligned}
e'_v &= h_L W^T + g(h_L) e_{aggregate} W^T ,\\
&=h_L W^T + \Delta(h_L) W^T
\end{aligned}
\end{equation}
\vspace{0.3em}

where $W$ is the weight matrix employed by the LLM. It should be noted that only the Steer Adapter and the Gating Network used to calculate $\Delta(h_L)$ contain parameters learnable during the training, while other parameters remain fixed. 

To minimize the learnable parameters in the Steer Adapter, we implement a parameterization approach inspired by low-rank adaptation (LoRA) \citep{hu2021loralowrankadaptationlarge}. Specifically, rather than learning a full-rank transformation matrix, we decompose it into two low-rank matrices, as shown in Figure \ref{fig:steer}. This decomposition maintains the transformation's expressive power while allowing efficient integration of personalized information, seamlessly merging it into the LLM's representations to facilitate effective user adaptation and stable generation. Additionally, it opens possibilities for incorporating more sophisticated personalized models into LLM generation.

\subsection{Additional Results}

\subsubsection{Effect of Inserted Context Length on Preference Consistency Error Types}
\label{Error_Types_Inserted Context Length}
\begin{figure*}[htbp]
\begin{center}
\includegraphics[width=0.9\textwidth,]
{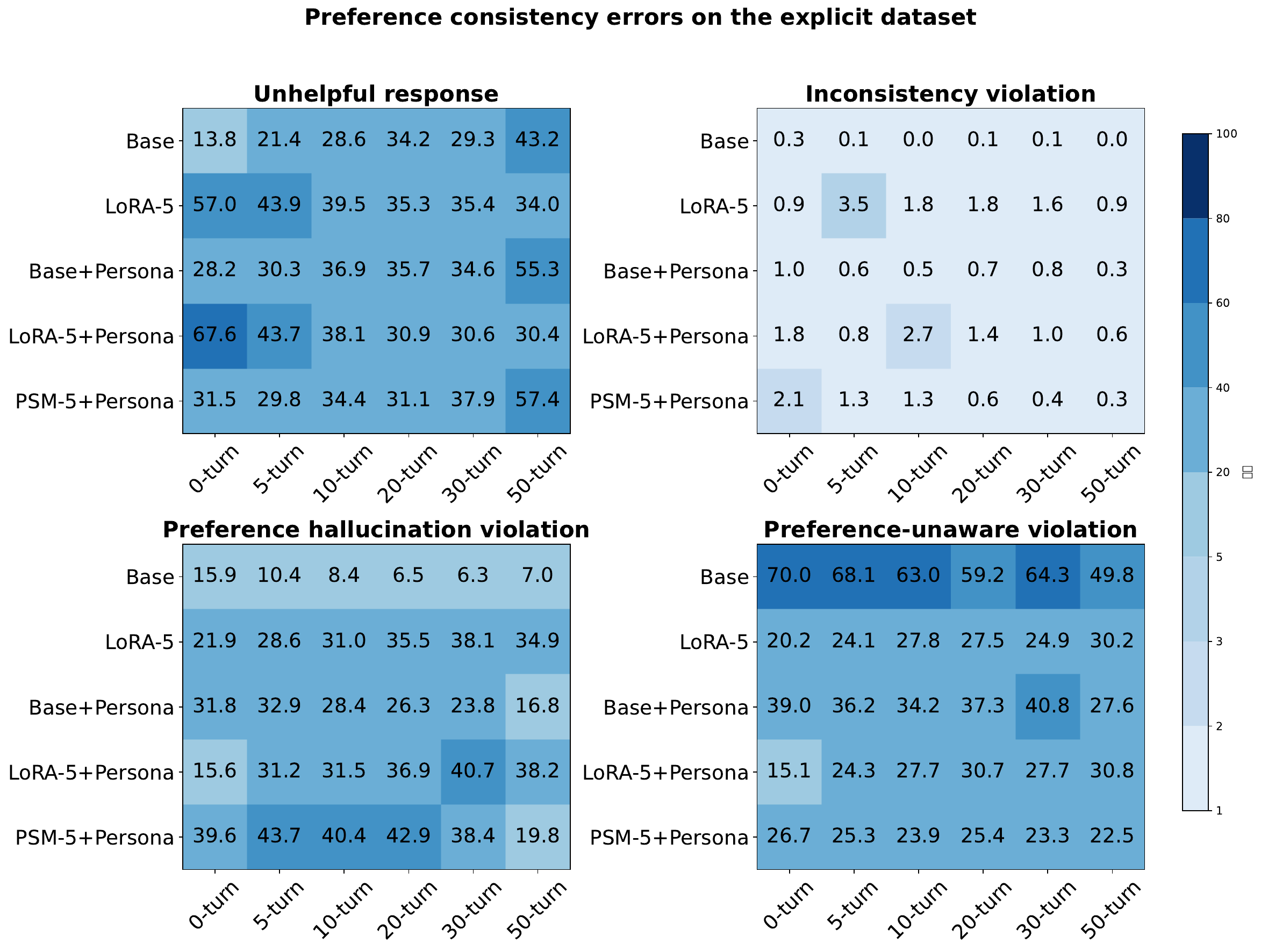} 
\end{center}
\caption{Preference consistency error types under different numbers of inserted irrelevant turns (n-turn).}
\label{fig:error_length_explicit}
\end{figure*}

\begin{figure*}[htbp]
\begin{center}
\includegraphics[width=0.9\textwidth,]
{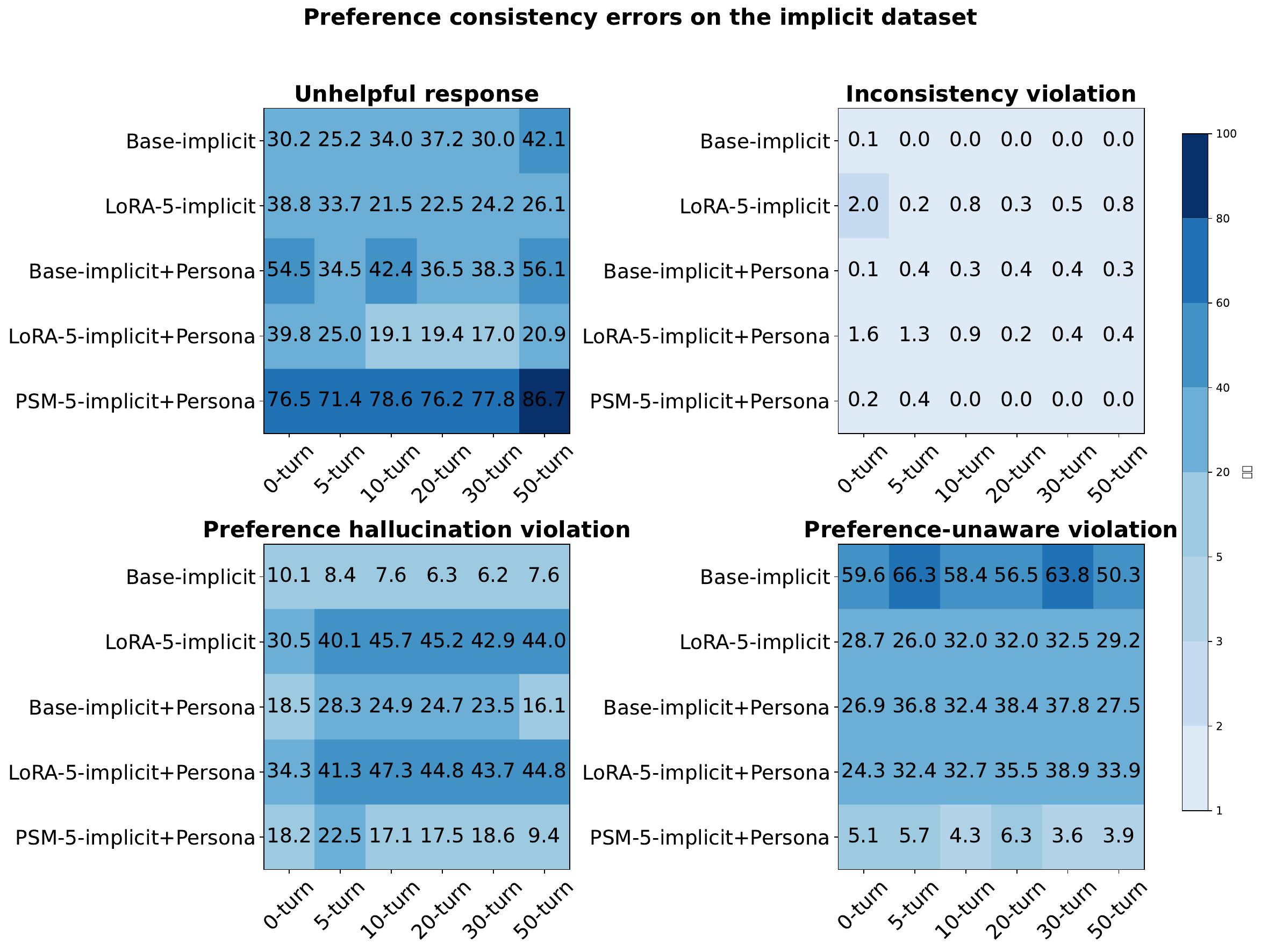} 
\end{center}
\caption{Preference consistency error types under different numbers of inserted irrelevant turns (n-turn).}
\label{fig:error_length_implicit}
\end{figure*}

Figures \ref{fig:error_length_explicit} and \ref{fig:error_length_implicit} illustrate the changes in preference consistency error types across different numbers of inserted irrelevant turns in the explicit and implicit datasets, respectively. Under zero-shot prompting without persona, Preference-Unaware Violations become increasingly prominent as the number of irrelevant turns increases, indicating that LLMs struggle more to maintain awareness of user preferences when exposed to longer irrelevant context. With the introduction of various methods, including finetuning approaches such as LoRA and PSM, the proportion of Preference-Unaware Violations decreases, while Hallucination Violations increase and Inconsistency Violations begin to appear, reflecting the challenges models face in generating responses that are both aligned with retrieved preferences and free from hallucinated information.

On the explicit dataset, LoRA is more prone to Inconsistency Violations across n-turn scenarios, whereas PSM exhibits higher rates of Preference Hallucination Violations in shorter contexts; however, as the number of irrelevant turns increases beyond 30, the rate of Preference Hallucination Violations in PSM decreases, while Unhelpful Responses become increasingly dominant. On the implicit dataset, Unhelpful Responses constitute the primary error type for PSM, indicating a tendency to refuse or provide unhelpful answers rather than attempt alignment with user preferences.

Overall, these results highlight the trade-offs between proactive preference-following and robustness to irrelevant context. Notably, as the length of irrelevant context increases, PSM becomes increasingly conservative, producing unhelpful responses, whereas LoRA is more proactive, continuing to attempt responses aligned with user preferences, although alignment issues remain.

\subsubsection{Effect of Inserted Context Length on Finetuning Results}
\label{EffectofInsertedctxft}
\begin{figure*}[htbp]
\begin{center}
\includegraphics[width=0.9\textwidth,]
{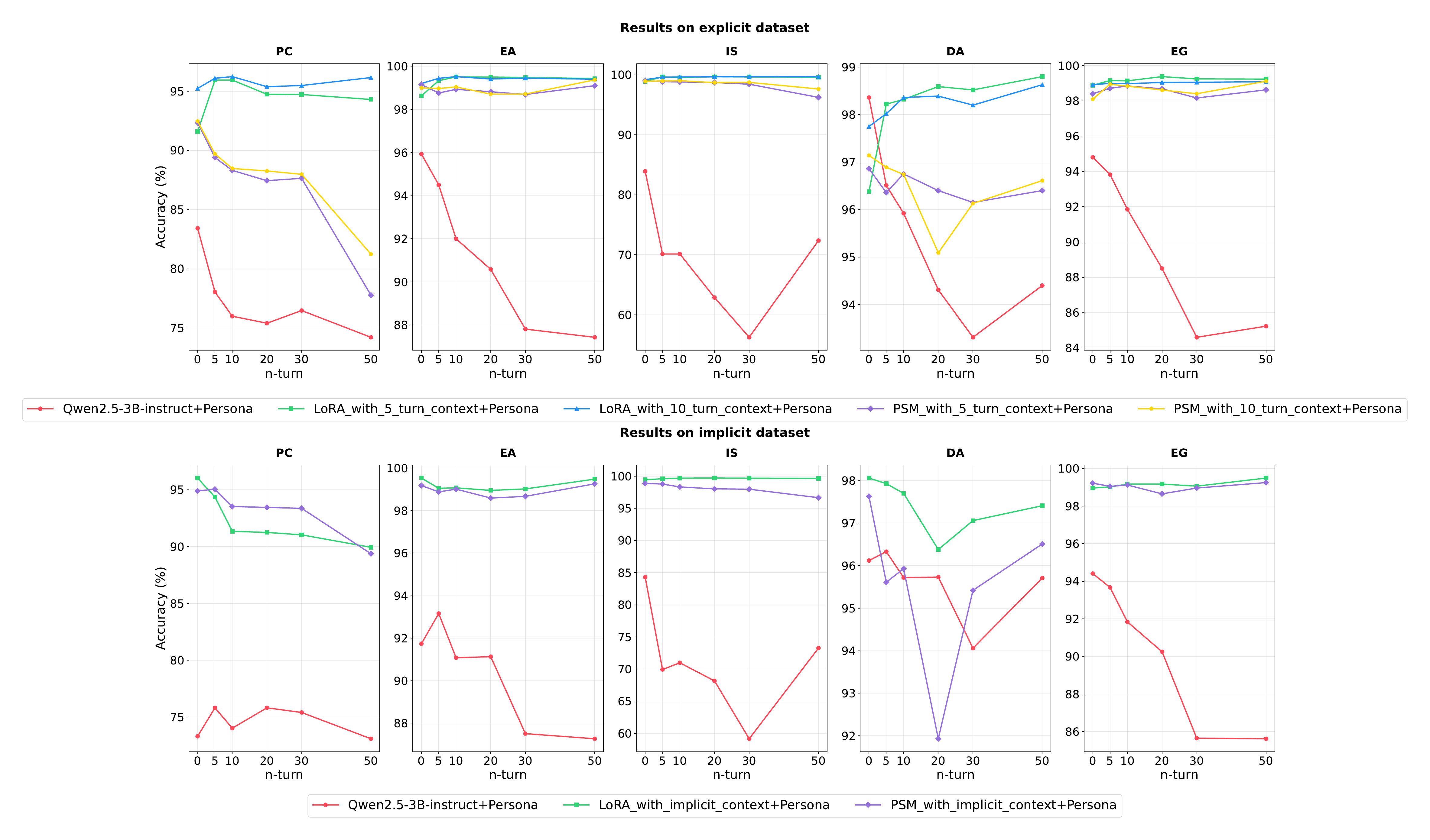} 
\end{center}
\caption{Accuracy of LLMs on preference consistency (PC) and child-oriented dimensions under different numbers of inserted irrelevant turns (n-turn).}
\label{fig:tuning}
\end{figure*}

Figure \ref{fig:tuning} illustrates how persona-informed finetuning methods (i.e., PSM and LoRA) evolve with increasing dialogue length on both explicit and implicit datasets (i.e., datasets with historical dialogues that explicitly and implicitly reveal the child preference). On the explicit dataset, PSM-based models show a relatively sharp decline in preference consistency as the number of inserted irrelevant dialogue turns increases, while LoRA-based models exhibit a moderate decrease. Moreover, results on the PC (Preference Consistency) dimension indicate that training with longer irrelevant dialogues yields greater robustness on equally long test dialogues than training with shorter ones. Interestingly, within the child-oriented dimensions, most metrics remain relatively stable across dialogue lengths, whereas developmental appropriateness (DA) exhibits the largest fluctuations, indicating its heightened sensitivity to contextual length.

On the implicit dataset, model trends largely mirror those observed on the explicit dataset. In the PC dimension, PSM-based models remain relatively stable compared to LoRA-based models on dialogues shorter than 30 turns (12K tokens) and benefit more from inserted irrelevant dialogues. However, in longer dialogues (e.g., 50 turns, PSM-based models show a sharper decline, falling below LoRA-based models. This pattern may be due to the fact that PSM relies on the aggregation of final-layer vectors to incorporate the persona information into an LLM, which works well when the inserted irrelevant dialogue is short, but may be negatively affected by accumulated noise when the irrelevant dialogue is long. In contrast, LoRA's low-rank adaptation maintains greater stability in extended contexts.

\subsubsection{Performance on English version}
\begin{figure*}[htbp]
\begin{center}
\includegraphics[width=0.9\textwidth,]
{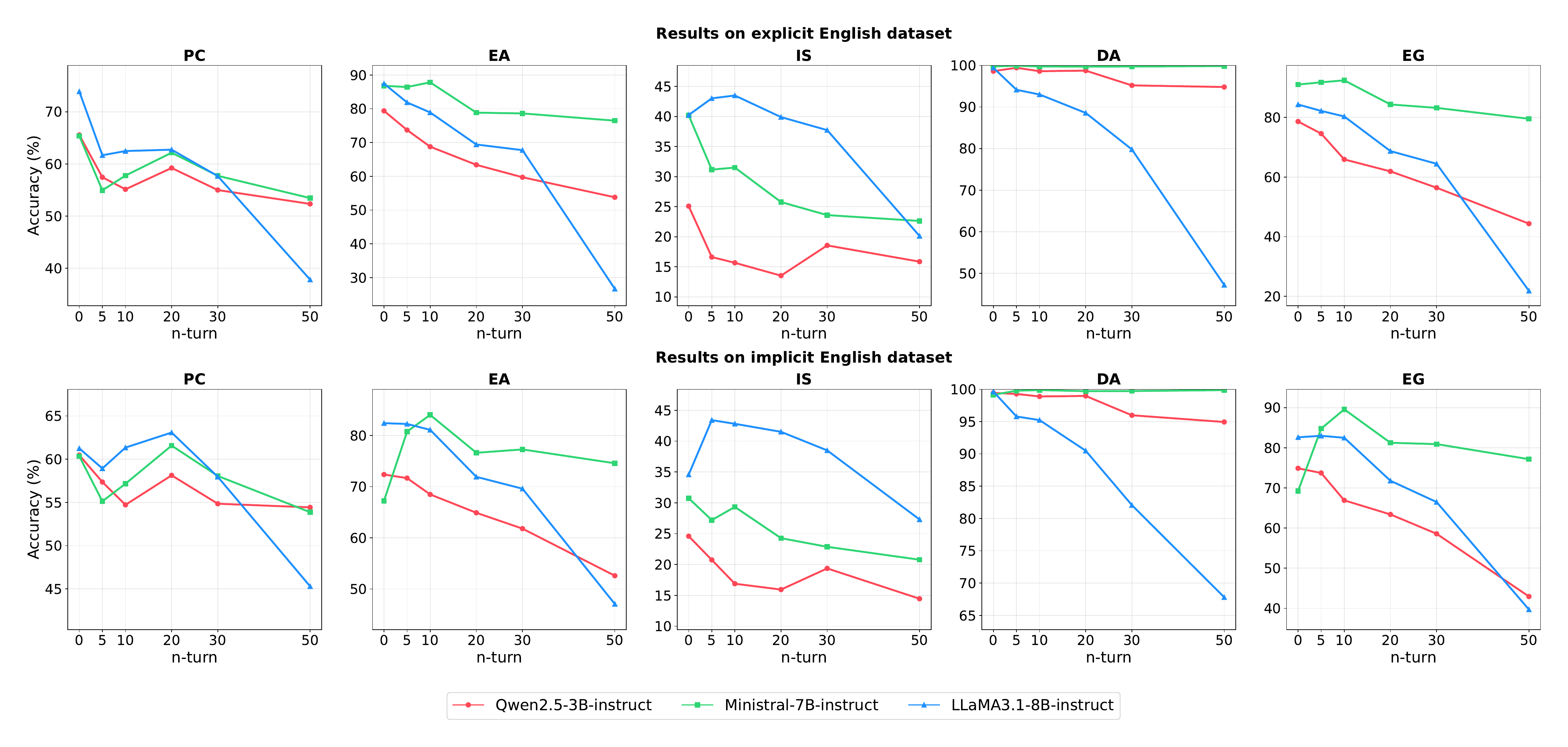} 
\end{center}
\caption{LLMs performances on preference consistency (PC) and the child-oriented evaluation under different numbers of inserted irrelevant dialogue turns on the English dataset.}

\label{fig:english_result}
\end{figure*}

To gain a comprehensive understanding, we conduct more experiments on the English version of ChildEval. Figure \ref{fig:english_result} presents the evaluation results across different numbers of inserted irrelevant dialogue turns. These dynamics indicate that tasks and models exhibit varying levels of robustness and adaptability across different dialogue stages in both the explicit and implicit datasets. IS remains the most challenging dimension for all models. However, the overall performance on the English dataset is slightly lower, likely because it is a translated counterpart of the Chinese corpus and may not fully capture the natural distribution of native English dialogues.

\begin{figure*}[htbp]
\begin{center}
\includegraphics[width=0.9\textwidth,]
{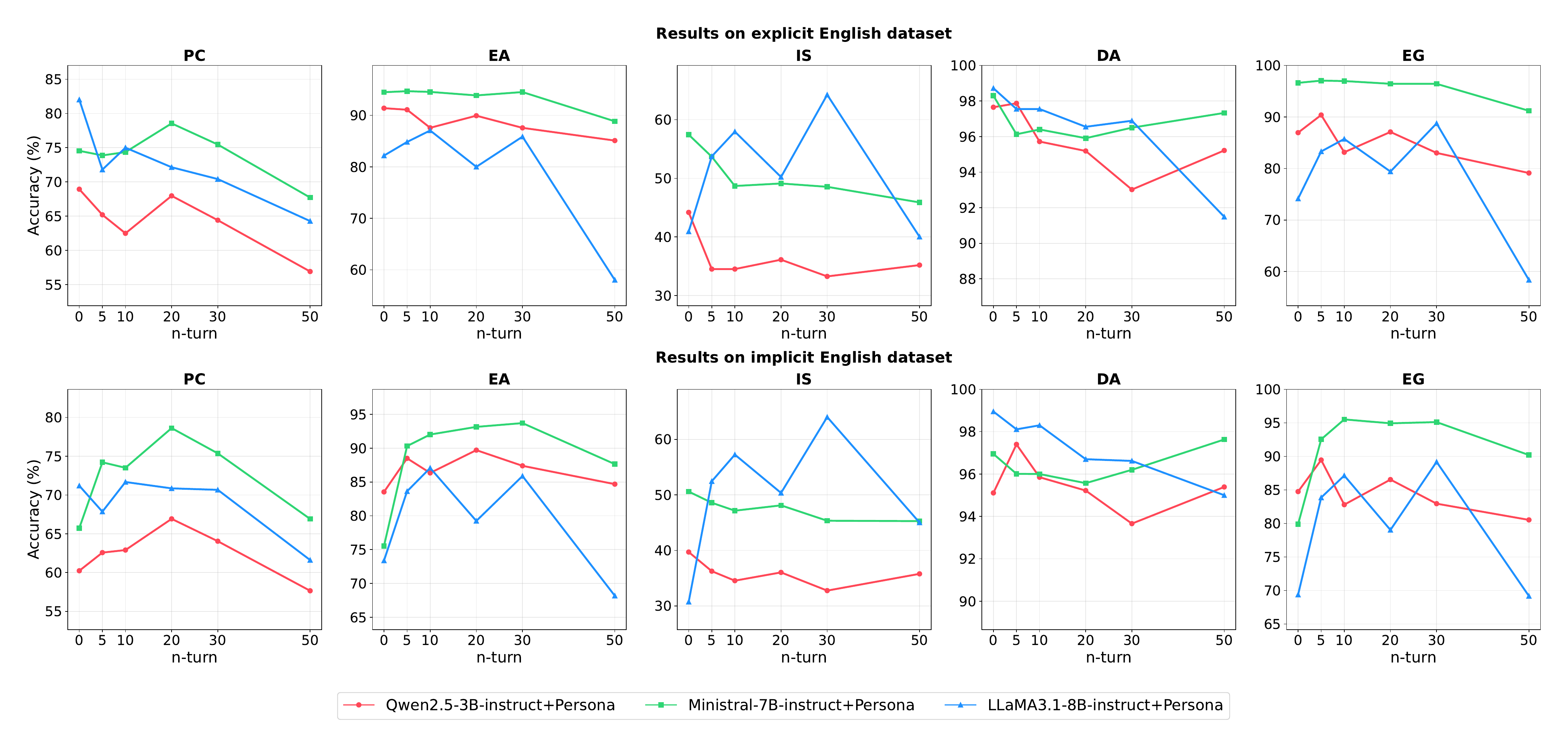} 
\end{center}
\caption{LLMs performances on preference consistency (PC) and the child-oriented evaluation under different numbers of inserted irrelevant dialogue turns on the English dataset, after integrating persona information into the prompt.}
\label{fig:english_result_persona}
\end{figure*}

Figure \ref{fig:english_result} presents the evaluation results after incorporating the persona information into the prompt, and different models exhibit divergent patterns. Notably, LLaMA3.1-8B-instruct shows substantial fluctuations on the EA, IS, and EG dimensions of the child-oriented evaluation. The performances of the other two models show a decreasing trend with small fluctuations as the number of irrelevant dialogue turns increases. Comparing Figure \ref{fig:english_result} and Figure \ref{fig:english_result_persona}, the inclusion of persona leads to significant improvements across PC, EA, IS, and EG for all models, with the only exception being a slight decrease on DA observed for Ministral-7B-instruct.

\label{eng_res}
\section*{Acknowledgments}
We appreciate the assistance provided by DeepSeek-R1 \citep{guo2025deepseek} , ChatGPT \citep{openai2024openaio1card} in writing aid and sentence-level polishing.

We would like to thank the volunteers for sharing anonymized and consented
AI-assisted interaction summaries involving their children (aged 3-6),
which were used solely to inform the design of child personas. Our thanks also go to the colleagues and the volunteers who reviewed the corpus and offered valuable suggestions. Finally, we are grateful to the anonymous reviewers and the area chair for their constructive feedback.
\end{document}